\documentclass{article} 

\usepackage[acronym,nowarn,section,nogroupskip,nonumberlist]{glossaries}

\glsdisablehyper{}
\newcommand{\acaps}[1]{{\scshape #1}}
\newacronym[\glslongpluralkey={Conditional Neural Processes}]{cnp}{\acaps{cnp}}{Conditional Neural Process}

\newacronym{VI}{vi}{variational inference}
\newacronym{KL}{kl}{Kullback-Leibler}
\newacronym{ELBO}{elbo}{\emph{evidence lower bound}}
\newacronym{MCMC}{mcmc}{Markov chain Monte Carlo}

\newacronym[\glslongpluralkey={Projection-based Equivariance Regularizer}]{per}{\acaps{per}}{Projection-based Equivariance Regularizer}

\newacronym{de}{\textsc{de}}{Deep Ensemble}
\newacronym{bma}{\textsc{bma}}{Bayesian Model Averaging}
\newacronym{mlp}{\textsc{mlp}}{Multilayer perceptron}
\newacronym{cnn}{\textsc{cnn}}{Convolutional Neural Network}
\newacronym{swa}{\textsc{swa}}{Stochastic Weight Averaging}
\newacronym{sgd}{\textsc{sgd}}{Stochastic Gradient Descent}
\newacronym{kd}{\textsc{kd}}{Knowledge Distillation}

\newacronym{sb}{\textsc{sb}}{Schr\"odinger Bridge}
\newacronym{dsb}{\textsc{dsb}}{Diffusion Schr\"odinger Bridge}
\newacronym{i2sb}{\textsc{i}\textsuperscript{2}\textsc{sb}}{Image-to-Image Schr\"odinger Bridge}

\newacronym{dbn}{\textsc{dbn}}{Diffusion Bridge Network}
\newacronym{ed}{\textsc{ed}}{Ensemble Distillation}
\newacronym{bn}{\textsc{bn}}{Bridge Network}
\newacronym{end2}{\textsc{e}n\textsc{d}\textsuperscript{2}}{Ensemble Distribution Distillation}

\newacronym{nfe}{\textsc{nfe}}{Number of Function Evaluations}
\newacronym{flops}{\textsc{flop}s}{FLoating point OPerations}
\newacronym{sde}{\textsc{sde}}{Stochastic Differential Equation}
\newacronym{dsc}{\textsc{dsc}}{Depthwise Separable Convolution}
\newacronym{ema}{\textsc{ema}}{Exponential Moving Average}
\usepackage{amsmath, amssymb, amsthm}
\usepackage{mathtools}
\usepackage{bbm}
\usepackage{dsfont}

\newcommand{\bp}{\mathbf{p}}

\newcommand{\bz}{\mathbf{z}}

\newcommand{\bZ}{\mathbf{Z}}

\newcommand{\bsh}{\boldsymbol{h}}

\newcommand{\bsp}{\boldsymbol{p}}

\newcommand{\bsx}{\boldsymbol{x}}
\newcommand{\bsy}{\boldsymbol{y}}
\newcommand{\bsz}{\boldsymbol{z}}

\newcommand{\bsZ}{\boldsymbol{Z}}

\newcommand{\calD}{{\mathcal{D}}}

\newcommand{\calL}{{\mathcal{L}}}

\newcommand{\calN}{{\mathcal{N}}}

\newcommand{\calT}{{\mathcal{T}}}
\newcommand{\calU}{{\mathcal{U}}}

\newcommand{\bbE}{\mathbb{E}}

\newcommand{\bbN}{\mathbb{N}}

\newcommand{\bbR}{\mathbb{R}}

\newcommand{\btheta}{{\boldsymbol{\theta}}}

\newcommand{\blambda}{{\boldsymbol{\lambda}}}

\newcommand{\bxi}{{\boldsymbol{\xi}}}

\newcommand{\bphi}{{\boldsymbol{\phi}}}

\newcommand{\bpsi}{{\boldsymbol{\psi}}}

\newcommand{\bTheta}{\boldsymbol{\Theta}}

\theoremstyle{plain}

\theoremstyle{definition}

\theoremstyle{remark}

\newcommand{\dee}{\mathrm{d}}
\newcommand{\grad}{\nabla}

\DeclareMathOperator*{\argmax}{arg\,max}

\newcommand{\normal}{\mathcal{N}}

\newcommand{\given}{\,|\,}

\def\[#1\]{\begin{equation}\begin{aligned}#1\end{aligned}\end{equation}}

\usepackage[usenames,dvipsnames]{xcolor}

\usepackage{graphicx}

\usepackage{booktabs, array}

\definecolor{citecolor}{RGB}{0,102,204}
\definecolor{linkcolor}{RGB}{190,105,30}
\definecolor{urlcolor}{RGB}{199,21,133}

\usepackage[colorlinks, linktoc=all,pagebackref=true]{hyperref}
\usepackage[all]{hypcap}
\hypersetup{citecolor=citecolor}
\hypersetup{linkcolor=linkcolor}
\hypersetup{urlcolor=urlcolor}
\usepackage[nameinlink,capitalise, noabbrev]{cleveref}
\creflabelformat{equation}{#2\textup{#1}#3}  
\crefname{section}{\S}{\S\S}

\usepackage{booktabs}
\newsavebox\CBox

\newcommand{\tbf}[1]{\sbox\CBox{#1}\resizebox{\wd\CBox}{\ht\CBox}{\textbf{#1}}}
\newcommand{\tpm}[1]{\small{$\pm{\, #1}$}}
\newcommand{\umet}[1]{#1 ($\uparrow$)}
\newcommand{\dmet}[1]{#1 ($\downarrow$)}
\newcommand{\relv}[1]{$\times$ #1}

\usepackage{ifthen}
\newcommand{\tv}[3][n]{%
    \ifthenelse{\equal{#1}{n}}{%
        \ifthenelse{\equal{#3}{}}{#2}{#2 \tpm{#3}}%
    }{\ifthenelse{\equal{#1}{b}}{%
        \ifthenelse{\equal{#3}{}}{\tbf{#2}}{\tbf{#2} \tpm{#3}}%
    }{%
        \textcolor{red}{ERROR}%
    }%
}}

\usepackage{algorithm}
\usepackage{algorithmic}

\usepackage{iclr2024_conference,times}

\usepackage{hyperref}
\usepackage{url}

\title{Fast Ensembling with Diffusion Schr\"odinger Bridge}

\author{Hyunsu Kim$^*$, Jongmin Yoon$^*$, Juho Lee\\
  Kim Jaechul Graduate School of AI\\
  KAIST\\
  Daejeon, South Korea \\
  \texttt{\{kim.hyunsu,jm.yoon,juholee\}@kaist.ac.kr} \\
}

\iclrfinalcopy 
\begin{document}

\maketitle

\begin{abstract}
\gls{de} approach is a straightforward technique used to enhance the performance of deep neural networks by training them from different initial points, converging towards various local optima. However, a limitation of this methodology lies in its high computational overhead for inference, arising from the necessity to store numerous learned parameters and execute individual forward passes for each parameter during the inference stage.
We propose a novel approach called~\gls{dbn} to address this challenge. Based on the theory of the Schr\"odinger bridge, this method directly learns to simulate an \gls{sde} that connects the output distribution of a single ensemble member to the output distribution of the ensembled model, allowing us to obtain ensemble prediction without having to invoke forward pass through all the ensemble models. 
By substituting the heavy ensembles with this lightweight neural network constructing \gls{dbn}, we achieved inference with reduced computational cost while maintaining accuracy and uncertainty scores on benchmark datasets such as CIFAR-10, CIFAR-100, and TinyImageNet. Our implementation is available at \url{https://github.com/kim-hyunsu/dbn}.


\end{abstract}

\glsresetall

\section{Introduction}
\label{main:sec:introduction}

\gls{de} \citep{lakshminarayanan2017simple} is one of the straightforward yet powerful techniques for improving the performance of deep neural networks. This method involves averaging the outputs of multiple models trained independently with different random initializations and data scan orders. \gls{de} is an instance of bagging~\citep{breiman96} where the models are trained with fixed datasets, and can be interpreted as an approximatation of \gls{bma}. \gls{de} can significantly improve the prediction accuracy of deep neural networks, and more importantly, the uncertainty quantification and out-of-distribution robustness for tasks across various domain. 

However, the major drawback of \gls{de} lies in the fact that the computational overhead and memory usage during inference calculations scale linearly with the number of ensemble members, which may be problematic for resource-limited environments or when a model is prohibitly large. Various approaches have been proposed to mitigate this issue, ideally reducing the inference cost of \gls{de} down to a single forward pass while minimizing the degradation in the predictive performance. 
A popular method in this direction is ensemble distillation, which is based on knowledge distillation~\citep{hinton2015distilling} over the ensemble outputs. In ensemble distillation, the average output of multiple ensemble members is set as the output of the teacher network, and a single model is set as a student whose output is learned to minimize the discrepancy from the teacher outputs.
Additionally, techniques involving the use of multi-head models \citep{tran2021hydra}, shared weights \citep{wen2019batchensemble}, learning ensemble distributions \citep{malinin2020ensemble}, employing diversity-promoting augmentation \citep{nam2021diversity}, or training generators simulating the ensemble predictions \citep{penso2022functional} have been presented.
However, these methods either still demonstrate performance inferior to \gls{de} or require an inference cost comparable to \gls{de} in order to achieve equivalent performance.

Recently, \citet{yun2023traversing} proposed an orthogonal approach to reduce inference costs of \gls{de}. They achieve this by connecting ensemble members through low-loss subspaces, employing techniques such as Bezier curves between each pair of ensemble members based on methods outlined by \citet{garipov2018loss}. For each ensemble member, a low-loss subspace is chosen, originating from that member, and a parameter within the subspace is sampled (usually from the center of the subspace). To streamline the inference process, they introduce \gls{bn}, a lightweight neural network that takes an intermediate feature from the ensemble member and directly predicts the output originally computed from the model on the low-loss subspace. The final output is approximated through an average of the ensemble member's output and the output predicted by \gls{bn}. Crucially, \gls{bn} is designed to have a negligible number of parameters and inference cost compared to the full ensemble model, resulting in significantly reduced inference costs. Experimental results using real-world image classification benchmarks and large-scale deep neural networks validate their approach, showcasing faster ensemble inference with sub-linear scaling of inference costs in relation to the number of ensemble members.

However, \gls{bn} presented in \citet{yun2023traversing} comes with critical limitations. First, it does not directly predicts the output from the other ensemble members, but only for the models on the low-loss subspaces. This incurs an extra training cost of learning low-loss subspaces between all pairs of ensemble members. Second, a single \gls{bn} can only be constructed between a pair of ensemble members. This means that as the number of ensemble members grow, the number of \glspl{bn} to be constructed grows quadratically. Due to this structure, even though they achieved sub-linear growth in inference cost, there still is a large room for improvement.

In this paper, improving upon \gls{bn}, we present a novel method for reducing inference costs of ensemble models. 
Given a set of ensemble members, we designate one of the ensemble members as a starting point. Then we train a mapping that transports an output from the starting point to the output computed from the ensemble model (averaged output). Compared to \gls{bn}, our approach does not require learning low-loss subspaces, and do not require learning as many mappings as the number of pairs among ensemble members, at the cost of increased complexity to learn the mapping. Instead it learns this mapping via \gls{dsb}~\citep{bortoli2021diffusion,liu23i2sb}, a powerful model that can build a stochastic path between two probability distributions.  Leveraging \gls{dsb}, we create a sequence of predictions that progressively transition from the predictions of the starting point model to those of the ensemble model, capturing the inherent correlations among these predictions. Importantly, recognizing our primary objective of lowering inference costs, we design the score network of \gls{dsb} to be lightweighted and incorporate diffusion step distillation to further minimize computational overhead. We refer to our approach as \gls{dbn}, and their experimental results demonstrate significant improvements in the efficiency of ensemble inference cost reduction, surpassing the performance of \gls{bn} and other ensemble distillation techniques.

\section{Backgrounds}
\label{main:sec:backgrounds}

\subsection{Deep Ensemble and Bridge Network}
In the ensemble methods as \gls{de} \citep{lakshminarayanan2017simple},  a single neural network is trained $M$ times with the same data but different random seeds (thus with different initializations and data processing orders), yielding $M$ different models with parameters $\{\btheta_i\}_{i=1}^M$ located in different modes in the loss surface. For the prediction, the outputs from those $M$ models are averaged at the output level to construct a final output. The $M$ members participating in ensemble often disagree on their predictions, thus giving functional diversity facilitating more accurate, robust, and better calibrated decision makings.
However, it requires $M$ number of models loaded on memory and $M$ number of forward computations.

The \gls{bn}~\citep{yun2023traversing} is one of methods suggested for reducing computational costs of ensemble methods. 
\gls{bn} hinges on the mode connectivity~\citep{garipov2018loss} of ensemble members, meaning that it is possible to connect two different modes via a low-loss subspace, indicating the intrinsic connection between them. The main intuition behind \gls{bn} is that, if we can learn such a low-loss subspace, then we may directly predict the outputs computed from the parameters on the subspace in the output level. In detail, consider a neural network $f_\btheta(\cdot)$ with a parameter $\btheta$ trained with a loss function $\calL(\btheta)$. Let $\btheta_i$ and $\btheta_j$ be two modes. \gls{bn} first search for a new parameter $\btheta_{i,j}(\alpha)$ that satisfies on the parametric Bezier curve,
\[
\btheta_{i,j}(\alpha) = (1-\alpha)^2\btheta_i + 2\alpha(1-\alpha)\btheta_{i,j} + \alpha^2\btheta_j
\]
where the anchor parameter $\btheta_{i,j}$ is obtained by minimizing
$
\bbE_{\alpha \sim \calU(0, 1)} \left[
    \calL\left(\btheta_{i,j}(\alpha) \right)
\right]$, so that the parameters $\{\btheta_{i,j}(\alpha)\}_{\alpha\in[0,1]}$ on the curve locate on the low-loss subspace. After building such a curve, a light-weight neural network $s$ (\gls{bn}) is trained, where $s$ gets a feature vector $\bsz_i$ of an input $\bsx$ computed solely from the first model $\btheta_i$ and try to predict the output evaluated at the center of the Bezier curve $\btheta_{i,j}(0.5)$ for the same input $\bsx$. That is, the \gls{bn} is an estimator trying to approximate $f_{{\btheta_{i,j}(0.5)}}(\bsx) \approx s(\bsz_i)$.
Then the ensemble of two models, $\btheta_i$ and $\btheta_{i,j}(0.5)$, can be approximated via the \gls{bn} as follows:
\[
\frac{1}{2}\bigg( f_{\btheta_i}(\bsx) + f_{\btheta_{i,j}(0.5)}(\bsx)\bigg) \approx \frac{1}{2}\bigg( f_{\btheta_i}(\bsx) + s(\bsz_i)\bigg).
\]
Since the $\btheta_{i,j}(0.5)$ encompasses the information of both $\btheta_i$ and $\btheta_j$, $s$ is expected to approximate the outputs to decent quality.  However, since \gls{bn} does not directly predict the output of $\btheta_j$ but only approximates the models with $\btheta_{i,j}(0.5)$, it has limitation in mimicking the actual ensemble predictions. In addition, a single bridge network is constructed between only a pair of given ensemble members. As a result, the number of \glspl{bn} to be constructed may grow quadratically in the number of ensemble members, leading to extra training costs.


\subsection{Diffusion Schr\"odinger Bridge}
\gls{dsb}~\citep{bortoli2021diffusion,chen2023likelihood} is a conditional diffusion model that solves~\gls{sb} problem~\citep{sch1932}, an entropy-regularized optimal transport problem that finds the diffusion process between two distributions.
Even though the~\gls{sb} problem provides the finite-time solution on finding the probability path, it requires iterative optimization and inference procedure which is time-consuming, and has rarely demonstrated its practicality in deep learning models albeit its theoretical soundness.
Recently, \citet{liu23i2sb} proposed a tractable special case of \gls{dsb} called \gls{i2sb} for an application of image manipulation such as image restoration or super-resolution.
In their work, a tractable special case of \gls{dsb} is proposed, and has demonstrated its efficiency for real-world image datasets.
Due to its training stability incurred from training a single score network, we choose it out of several \glspl{dsb} despite of its strict condition.
In this section, we introduce a brief overview of~\gls{i2sb} method.

First of all, Schr\"odinger Bridge (SB) is an optimal transport problem that seeks to find the forward and backward processes
\[
\dee \bsZ_t &= [f_t+\beta_t\nabla\log \Psi(\bsZ_t, t)]\dee t + \sqrt{\beta_t}\dee W_t,\quad \bsZ_0\sim p_0\\
\dee \bsZ_t &= [f_t-\beta_t\nabla\log \hat{\Psi}(\bsZ_t, t)]\dee t + \sqrt{\beta_t}\dee \overline{W}_t,\quad \bsZ_1\sim p_1
\label{eq:sb}
\]
where $(p_0, p_1)$ are the boundary distributions, $\{W_t, \overline{W}_t\}$ are the standard Wiener process and its time-reversal, and $\{f_t, \beta_t\}$ are the drift and diffusion coefficients.
If the pair of functions $\{\Psi, \hat{\Psi}\}$ solves the following coupled PDE
\[
\frac{\partial \Psi(\bsz_t,t)}{\partial t}       = \grad\Psi^\top f_t - \frac{1}{2} \beta_t \Delta \Psi,\quad\frac{\partial \hat{\Psi}(\bsz_t,t)}{\partial t} = - \grad\cdot(\hat{\Psi} f_t) + \frac{1}{2} \beta_t \Delta \Psi\\
\text{with }
\Psi(\bsz_0, 0)\hat{\Psi}(\bsz_0, 0) = p_0(\bsz_0), \Psi(\bsz_1, 1)\hat{\Psi}(\bsz_1, 1) = p_1(\bsz_1),
\label{eq:fokker_planck}
\]
then~\eqref{eq:sb} provides the optimal solution to an entropy-regularizing optimization problem that finds the optimal path between $p_0$ and $p_1$.
Then~\eqref{eq:fokker_planck} and its time-reversal directly follows the Fokker-Planck equation of the SDE in~\eqref{eq:sde_forward} as follows, respectively.
\[\label{eq:sde_forward}
\dee \bsZ_t=f_t \dee t + \sqrt{\beta_t}\dee W_t,\quad \bsZ_0\sim\hat{\Psi}(\cdot,0)\text{ and }
\dee \bsZ_t=f_t \dee t + \sqrt{\beta_t}\dee \overline{W}_t,\quad \bsZ_1\sim\Psi(\cdot,1).
\]
However, both $\Psi$ and $\hat{\Psi}$ are intractable drifts so~\gls{i2sb} assumes a certain form of the boundary distributions $p_0$ and $p_1$.
Followed from~\citet{liu23i2sb}, we take the energy potential functions $\widehat{\Psi}(\cdot,0)=p_0(\cdot):=\delta_a(\cdot)$ and $\Psi(\cdot,1)=p_1(\cdot)/\widehat{\Psi}(\cdot,1)$.
Here $\delta_a(\cdot)$ is the Dirac delta distribution centered at $a\in\bbR^d$ which makes the diffusion process computationally tractable.
Then, we can approximate both forward and backward \glspl{sde} using a single score network in the framework of DDPM \citep{ho2020ddpm} with the following Gaussian posterior:
\[\label{eq:i2sb_conditional}
\bsZ_t\sim q(\bsZ_t\given \bsZ_0,\bsZ_1)=\normal(\bsZ_t;\mu_t,\Sigma_t),\,\,
\mu_t=\frac{\overline{\sigma}_t^2}{\overline{\sigma}_t^2+\sigma_t^2}\bsZ_0+\frac{\sigma_t^2}{\overline{\sigma}_t^2+\sigma_t^2}\bsZ_1, \,\,\Sigma_t = \frac{\overline{\sigma}_t^2\sigma_t^2}{\overline{\sigma}_t^2+\sigma_t^2}, 
\]
where $\sigma_t^2=\int_0^t\beta_{t'}\dee t'$ and $\overline{\sigma}_t^2=\int_t^1\beta_{t'}\dee t'$ are cumulative forward and backward noise variances, respectively.
For algorithmic design for restoration problems, we take $p(\bsZ_0,\bsZ_1)=p_0(\bsZ_0)p_1(\bsZ_1|\bsZ_0)$ and $f=0$, and construct tractable~\glspl{sb} between individual samples from $\bsZ_0$ and $p_1(\bsZ_1|\bsZ_0)$.

\section{Diffusion Bridge Networks}
\label{main:sec:methods}

\begin{figure*}[t]
    \centering
    \includegraphics[width=0.65\textwidth]{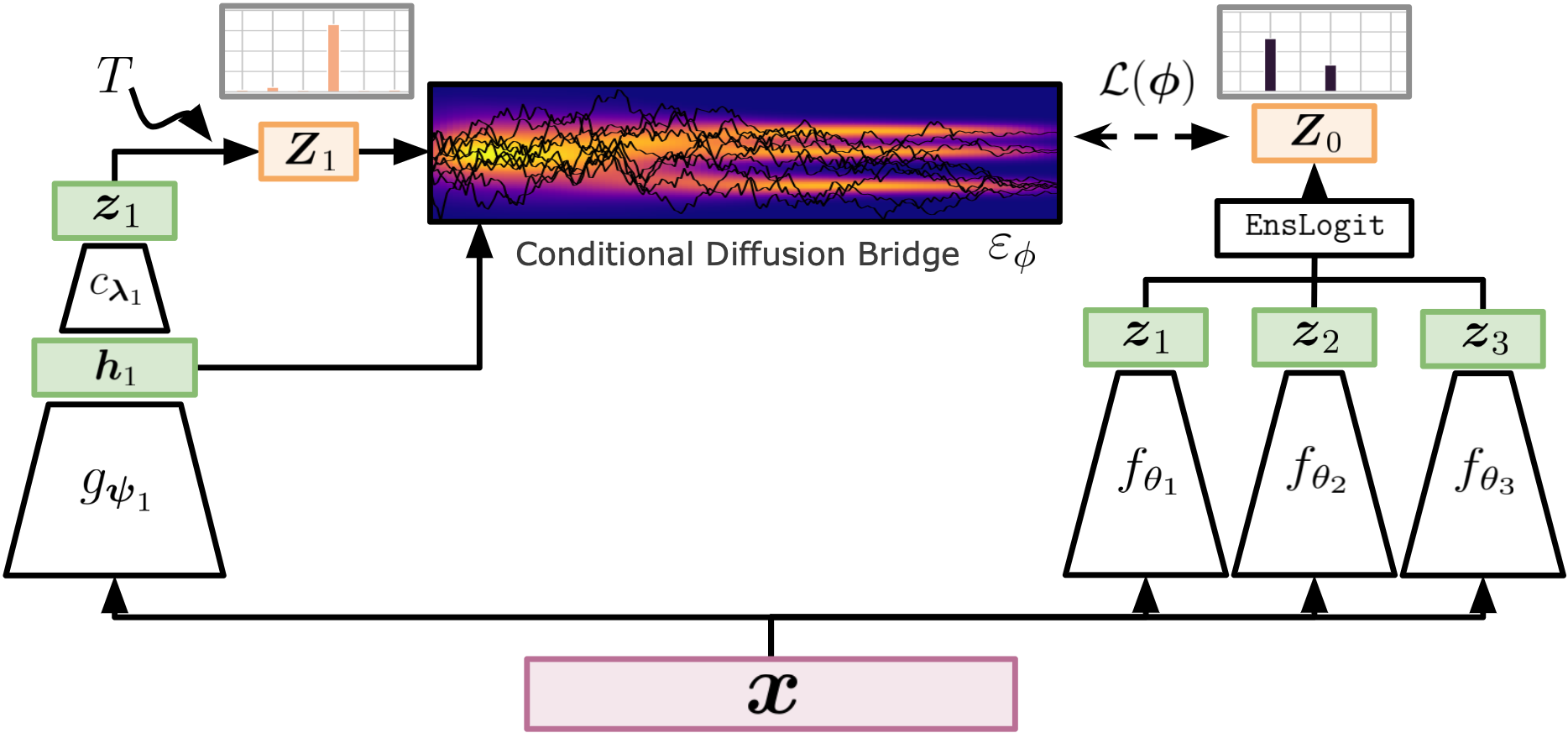}
    \vspace{-3mm}
    \caption{Overview of \gls{dbn}. For a given data, the conditional diffusion bridge learns a transition between logit distribution of one of the ensembles (left; source) and that of the target ensemble models (right; target).}
    \label{fig:concept}
\end{figure*}

In the previous \gls{bn} \citep{yun2023traversing}, there was a limitation in learning the transport between independent models, prompting them to adopt an alternative approach: it utilizes a neural network to predict the output of local optima aligned between the source and target models in terms of a low loss subspace instead of directly predicting the output of the target model. To overcome this limitation, instead of tackling the problem of predicting the ensemble outputs as one-step prediction (a single neural network evaluation), we cast the problem as a diffusion bridge construction between the output distributions of ensemble members.

\subsection{Settings and Notations}
We restrict our focus on the $K$-way classification problem, where the goal is to train a classifier taking $d$-dimensional inputs and predict $K$-dimensional classification logits.
We assume that a classifier with parameter $\btheta$ is decomposed into two parts, a feature extractor $g_{\bpsi}: \bbR^d \to \bbR^h$ that first encodes an input $\bsx$ into a feature vector $\bsh$, and then a classifier $c_{\blambda}: \bbR^h \to \bbR^K$ that transforms the feature vector $\bsh$ into a logit $\bsz$ of the class probability. That is, $f_{\btheta}(\bsx) = (c_{\blambda} \circ g_{\bpsi})(\bsx)$ and $\btheta = (\bpsi, \blambda)$. A collection of $M$ ensemble members are denoted as $\bTheta=\{\btheta_i\}_{i=1}^M$ and $\btheta_i = (\bpsi_i, \blambda_i)$.

\subsection{Conditional diffusion bridge In logit space}\label{sec:main}
Our goal is to construct a conditional diffusion bridge on the logit space, that is, a stochastic path $\{ \bsZ_t \}_{t \in [0,1]}$ where $\bsZ_1$ is constructed from the logit of the source model and $\bsZ_0$ from that of the target ensemble model. More specifically, given $\bTheta$, we choose one of the ensemble members $\btheta_1$ as a source model, and set $\bsz_1$ be the logit distribution computed from $\btheta_1$. Then the target $\bsz_0$ is set to be the logit distribution of the ensembled prediction.

More specifically, let $\bsx \in \bbR^d$ be an input, and let $\bsh_1 = g_{\bpsi_1}(\bsx)$ be the corresponding feature vector computed from the source model $\btheta_1$. Conditioned on $\bsx$, we define the source logit distribution as an implicitly defined distribution as follows,
\[\label{eq:i2sb_source_dist}
\bsz_1 = c_{\blambda_1}(\bsh_1), \quad T \sim p_\text{temp}, \quad \bsZ_1 = \frac{\bsz_1}{T},
\]
where $T$ is a annealing temperature drawn from some distribution $p_\text{temp}$.
The target logit, $\bsZ_0$, is then set to be the logit computed by the ensemble model as follows, for $\bsz^{(i)} := f_{\btheta_i}(\bsx)$,
\[
 \bp_i = \mathtt{Softmax}(\bsz^{(i)}), \,\,
 \bsZ_0 = \mathtt{EnsLogit}( \{ \bsz^{(i)} \}_{i=1}^M ) := \log \bar\bp - \frac{1}{K} \sum_{k=1}^K \log \bar\bp_{i, k},
\]
where the ensemble output $\bar{\bp} = \sum_{i=1}^M\bp_i/M$. Then, we get $\mathtt{Softmax}(\bsZ_0) = \bar\bp$, allowing us to get ensemble output $\bar\bp$ through approximating $\bsZ_0$. Next we construct \gls{i2sb} between $\bsZ_1$ and $\bsZ_0$.

The intuition behind this construction is as follows. If we directly use the original logit $\bsz_1$, the learned \gls{dsb} can easily be trapped in a trivial solution where it just produces the copy of the source logit $\bsz_1$ along the path $\{\bsZ_t\}_{t\in [0,1]}$, as the discrepancy between the source logit and the target logit is not large compared to the typical situation for which \gls{dsb} is constructed. That is, the source $\bsz_1$ can be a strong hint that acts a simplicity bias for \gls{dsb} learning. Also, \gls{i2sb} requires the source of the diffusion to be a conditional probability distribution, but $\bsz_1$ is a deterministic value. Hence by randomly annealing the source logit via a temperature $T$, we can naturally construct the source as the distribution of the annealed logits and also dilute the information included in $\bsz_1$, and this encourages \gls{dsb} to discover non-trivial paths between the source and the target. 

Based on the formulation on constructing the conditional diffusion bridge between the source and target distributions, we build an approximate reverse \gls{sde} derived from~\eqref{eq:sb} that simulates the path from $\bsZ_1$ to $\bsZ_0$ by estimating the score function $\grad \log \hat{\Psi}(\bsZ_t, t|\bsh_1) = \varepsilon_\phi(\bsh_1,\bsZ_t,t)/\beta_t$ as
\[\label{eq:reverse_i2sb}
\dee\bZ_t = (\beta_t/\sigma_t)\varepsilon_{\bphi}(\bsh_1, \bsZ_t, t) \dee t + \sqrt{\beta_t} \dee W_t, \quad \bsZ_1 \sim p_1(\bsZ_1\given \bsx),
\]
where $p_1$ is the distribution of $\bsZ_1$ implicitly defined as in \eqref{eq:i2sb_source_dist} and
$\varepsilon_{\bphi}: \bbR^{h} \times \bbR^{K} \times [0,1] \to \bbR^K$ is the score function estimator built with a neural network parameterized by $\bphi$,
and is trained to estimate the score function $\nabla_{\bsZ} \log p_t(\bsz_t\given \bsx)$ by minimizing the following objective function,
\[\label{eq:dbn_loss}
\calL(\bphi) = \bbE_{\bsx, \bsZ_t}
\left[ 
\left\Vert \varepsilon_{\bphi}(\bsh_1, \bsZ_t, t) - \frac{\bsZ_t - \bsZ_0}{\sigma_t}
\right\Vert_2^2
\right],
\]
with $\bsx \sim p_\text{data}$, $t \in \calU([0,1])$, and $\bsZ_t \sim q(\bsZ_t\given \bsZ_0, \bsZ_1)$ as defined in \eqref{eq:i2sb_conditional}.
The overall training pipeline is summarized in \cref{alg:training}.

\begin{algorithm}[t]
\small
    \caption{Training \glspl{dbn}}
    \label{alg:training}
    \begin{algorithmic}
        \REQUIRE{An (empirical) data distribution $p_\text{data}$, a temperature distribution $p_\text{temp}$,\\ ensemble parameters $\{\btheta_i\}_{i=1}^M$, and the score network $\varepsilon_\bphi$.}
        \STATE Fix a source model $f_{\btheta_1}$.
        \WHILE{not converged}
            \STATE Sample $\bsx \sim p_\mathrm{data}$.
            \FOR{$i=1$ to $M$}
                \STATE Compute the logits $\bz_i=f_{\btheta_i}(\bsx)$.
            \ENDFOR
            \STATE Get a target ensemble logit $\bZ_0=\mathtt{EnsLogit}(\{ \bsz_i \}_{i=1}^M)$.
            \STATE Draw a temperarture $T \sim p_\text{temp}$ and compute the annealed source logit $\bsZ_1 = \bsz_1 / T$.            
            \STATE Compute the loss according to \eqref{eq:dbn_loss}, and update
            $\bphi\gets \bphi - \eta \nabla_\bphi\calL(\bphi)$.
        \ENDWHILE
        \STATE \textbf{return} $\bphi$.
    \end{algorithmic}
\end{algorithm}

\subsection{Distillation of Diffusion Bridge}
Even though the family of diffusion models~\citep{ho2020ddpm,bortoli2021diffusion,liu23i2sb} achieves superior performance in learning generative models or constructing paths between distributions, they typically suffer from the slow sampling speed due to a large number of function evaluations required for simulation. The distillation techniques for diffusion models, which distill multiple steps of the reverse diffusion process to a single step, do not significantly harm the generation performance while accelerating the sampling speed.
In this paper, we adapt the progressive distillation proposed in \citep{salimans2022progressive} to reduce the sampling cost of \gls{dbn}.

Let $\calT = \{t_i\}_{i=1}^N$ be the discretized time interval used for diffusion bridge, with $0=t_0<t_1<\dots<t_N=1$. Then let $\calT' := \{t_j' \}_{j=1}^{N'}$ be the distilled time interval 
with $\calT' \subset \calT$. Let $\bsZ_{t_{i-1}} \sim p_{\bphi}(\bsZ_{t_{i-1}}\given\bsZ_{t_i})$ be the sample from an ancestral sampling following the score network $\varepsilon_{\bphi}$. Then a distilled score network with parameter $\bphi'$ is then trained with the loss
\[\label{eq:distill_loss}
\calL_\text{distill}(\bphi') = 
\bbE_{\bsx,j}\left[\left\Vert
\varepsilon_{\bphi'}(\bsh_1, \bsZ_{t'_j}, t_j') - \frac{\bsZ_{t_j'} - \bsZ_{{t_{j-1}'}}}{\sigma_{t_j'}}
\right\Vert_2^2\right],
\]
where $j \sim \calU(\{1,\dots,N'\})$ and $\bsZ_{t'_{j-1}} \sim \prod_{s=i-k}^{i-1} p_{\bphi}(\bsZ_{t_s}\given \bsZ_{t_{s+1}})$ with $t_{j-1}'=t_{i-k}$ and $t_j'=t_i$. This distillation is then recursively repeated until there only remains a single time step $(N'=1)$.


\subsection{Inference Procedure}
The inference with \gls{dbn} consists of forwarding an input through the source model and computing a single diffusion step from the model distilled by \eqref{eq:distill_loss}.
Given an input $\bsx$, we first compute its feature $\bsh_1$ using the feature extractor of the source model $g_{\bpsi_1}$ with $\bsh_1=g_{\bpsi_1}(\bsx)$.
Then we initialize the diffusion by drawing $T \sim p_\text{temp}$ and put $\bsZ_1 = \bsz_1/T$. The corresponding ensembled prediction $\bsy$ is then approximated as,
\[\label{eq:dbn_inference}
&\bsZ_0 = \bsZ_1 + (\beta_1/\sigma_1)\varepsilon_{\bphi '}(\bsh_1, \bsZ_1, 0) + \bxi_1 \text{ where } \bxi_1 \sim \calN(0, \Sigma_1), \\
& p(\bsy\given \bsx, \{\btheta_i\}_{i=1}^M) = \frac{1}{M} \sum_{i=1}^M \mathtt{Softmax}(\bsz^{(i)}) 
\approx \mathtt{Softmax}(\bsZ_0).
\]

\subsection{Combining multiple \glspl{dbn}}
\label{main:sec:multiple}
Note that the size of the score network $\varepsilon_{\bphi'}$ should be limited, because otherwise the cost from the diffusion simulation can outnumber the cost of computing the full ensemble. Hence, there is an intrinsic limit in the capacity of $\varepsilon_{\bphi'}$ representing the path between the source model and an full ensembled model (we study this capacity empirically in~\cref{main:sec:capacity}). When a single \gls{dbn} reached out to the limit, we may introduce multiple \glspl{dbn}, increasing the approximation quality at the cost of additional inference time.  Given $M$ ensemble models, we first build a \gls{dbn} starting from a source model $\btheta_1$. Then we build another \gls{dbn} sharing the same model $\btheta_1$ as the source and so on. Let $\{ \varepsilon_{\bphi'}^{(\ell)}\}_{\ell=1}^L$ be a set of score networks built in that way, with $L < M$. Then we can approximate the ensembled prediction $\bsy$ for $\bsx$ as,
\[
p(\bsy\given \bsx, \{\btheta_i\}_{i=1}^M) \approx \frac{1}{L}\sum_{\ell=1}^L\mathtt{Softmax}\left( \bsZ_0^{(\ell)}\right),
\]
where $\bsZ_0^{(\ell)}$ is the sample drawn as in \eqref{eq:dbn_inference} with $\ell^{\text{th}}$ score network.  
In our implementation, we combine $L$~\glspl{dbn} where $l^{\text{th}}$~\gls{dbn} estimates the recovered predictions by a collection of $N+1$ models $\{f_{\btheta_1}, f_{\btheta_{(l-1)(N-1)+2}},\cdots,f_{\btheta_{(l-1)(N-1)+N+1}}\}$, approximating the ensembled prediction of $M = LN+1$ ensemble models. For example, to simulate total $M=5$ ensemble models $\{\btheta_1,\btheta_2,\btheta_3,\btheta_4,\btheta_5\}$, we first train a score net that estimate ensemble distribution of $\{\btheta_1,\btheta_2,\btheta_3\}$ starting from $\btheta_1$ and secondly train another score net approximating $\{\btheta_1,\btheta_4,\btheta_5\}$ starting from $\btheta_1$. Note that the source models for $L$ \glspl{dbn} need not be different; in \cref{main:sec:experiments}, we show that a \emph{single} source model can be shared for all multiple \glspl{dbn}, minimizing the additional inference cost while significantly improving the accuracy.

\section{Experiments}
\label{main:sec:experiments}

\paragraph{Settings.}
We evaluate our approach using three widely adopted image classification benchmark datasets: CIFAR-10, CIFAR-100, and TinyImageNet \citep{tinyimagenet}.
In our experiment, the ensemble models that we construct to serve as bridges adopt the configuration outlined in the Bridge Network \citep{yun2023traversing} and are trained based on the ResNet architecture.
We use widely used ResNet-32$\times2$, ResNet-32$\times4$, and ResNet-34 networks \citep{he2016deep} for baseline ensemble classifiers for CIFAR-10, CIFAR-100, and TinyImageNet datasets, respectively.
In this context, the suffices "$\times2$" and "$\times4$" denote that the channel widths of the convolutional layers are multiplied by 2 and 4 from the conventional ResNet-32, respectively.
For the score network, inspired by \citet{sandler2018mobilenetv2}, we utilize residual connections \citep{he2016deep} and \gls{dsc}~\citep{chollet2017xception} to mitigate redundant computations and implement a light-weighted score network. Further details on datasets, model architectures, and hyperparameter settings used to evaluate our experiments are listed in~\cref{app:sec:experiment}.

\paragraph{Training.}
We train a single score network with 3 ensembles. If we need to mimic more than 4 ensembles, we can average 2 or more diffusion bridges according to~\cref{main:sec:multiple}. Since they share the source model, we can easily enhance the performance with low extra costs. We train the diffusion bridge with 5 steps before distillation for fast and efficient training of the score networks and 5 steps are enough to approximate the transport between the two conditional logit distributions.

\paragraph{Baseline methods.}
We validate how well our method accelerates the inference speed with lightweight networks by comparing our~\gls{dbn} with the target \gls{de}~\citep{lakshminarayanan2017simple} as the oracle ensemble.
We also compare our method with three existing methods that are widely used in fast and efficient ensemble distillation or bridge networks:
\gls{ed}~\citep{hinton2015distilling}, \gls{end2}~\citep{ryabinin2021scaling}, and \gls{bn} \citep{yun2023traversing}. We use more refined version of \gls{end2}  \citep{ryabinin2021scaling} instead of the original \gls{end2} \citep{malinin2020ensemble} that improved convergence.

\paragraph{Metrics.}
We measure the computational cost of each model in terms of \gls{flops} and the number of parameters (\#Params).
\gls{flops} count the number of additions and multiplications operations, and it represents a cost during an inference of a model.
\#Params represents the memory usage required for the inference. For the performance, we consider the classification accuracy (ACC), Negative Log-Likelihood (NLL), Brier Score (BS)~\citep{brier1950verification}, Expected Calibration Error (ECE)~\citep{guo2017on}, and Deep Ensemble Equivalent (DEE)~\citep{ashukha2020pitfalls}.
BS and ECE measure how much the classification output (confidence) is aligned with the true probability and thereby it implies the reliability of the model output.
DEE approximates the number of \gls{de} similar to a given model in terms of NLL. 
More profound formulations of the metrics are described in~\cref{app:sec:metrics}.

\subsection{Classification Performance and Uncertainty Metrics}
\begin{table*}[t]
    \newcommand{\metricrule}{\cmidrule(lr){2-3} \cmidrule(lr){4-8}}
    \newcommand{\modelrule}{\cmidrule(lr){1-8}}
    \renewcommand{\arraystretch}{1.18}
    \centering
    \caption{
        Performance on CIFAR-10/100, and TinyImageNet. \gls{bn}\textsubscript{medium} is the standard size of \gls{bn} and \gls{bn}\textsubscript{small} has reduced channels compared to \gls{bn}\textsubscript{medium}. The parentheses next to each model (e.g. (\gls{de}-3)) means the number of \glspl{de} that each model learns as a target.
    }
    \vspace{-2mm}
    \resizebox{0.95\textwidth}{!}{
        \begin{tabular}{lrrllllll}
            \\
            \textbf{CIFAR-10} \\
            \toprule
            Model                 & \dmet{FLOPs} & \dmet{\#Params} & \umet{ACC}          & \dmet{NLL}           & \dmet{BS}           & \dmet{ECE}            & \umet{DEE}           \\
            
            \midrule
            
            ResNet (\gls{de}-1)   & \relv{1.000} & \relv{1.000}    & \tv[n]{91.30}{00.10} & \tv[n]{0.3382}{0.0023} & \tv[n]{0.1409}{0.0011} & \tv[n]{0.0658}{0.0003} & \tv[n]{1.000}{}      \\

            \metricrule

            \quad +2 \gls{bn}\textsubscript{small} (\gls{de}-3)  & \relv{1.125} & \relv{1.097}    & \tv[n]{91.79}{00.05} & \tv[n]{0.2579}{0.0009} & \tv[n]{0.1198}{0.0003} & \tv[n]{0.0599}{0.0037} & \tv[n]{1.899}{}      \\

            \quad +4 \gls{bn}\textsubscript{small} (\gls{de}-5)  & \relv{1.245} & \relv{1.283}    & \tv[n]{91.87}{00.05} & \tv[n]{0.2580}{0.0010} & \tv[n]{0.1195}{0.0004} & \tv[n]{0.0624}{0.0098} & \tv[n]{1.898}{}      \\

            \quad +2 \gls{bn}\textsubscript{medium} (\gls{de}-3)  & \relv{1.411} & \relv{1.319}    & \tv[n]{91.91}{00.04} & \tv[n]{0.2544}{0.0011} & \tv[n]{0.1182}{0.0005} & \tv[b]{0.0591}{0.0096} & \tv[n]{1.938}{}      \\
            
            \quad +1 \gls{dbn} (\gls{de}-3)  & \relv{1.166} & \relv{1.213}    & \tv[b]{92.98}{00.16} & \tv[b]{0.2403}{0.0027} & \tv[b]{0.1084}{0.0013} & \tv[n]{0.0666}{0.0010} & \tv[b]{2.363}{}      \\

            \quad +2 \gls{dbn} (\gls{de}-5) & \relv{1.332} & \relv{1.426}    & \tv[b]{93.23}{00.07} & \tv[b]{0.2247}{0.0005} & \tv[b]{0.1033}{0.0005} & \tv[n]{0.0662}{0.0009} & \tv[b]{3.031}{}      \\

            \gls{ed} (\gls{de}-3)            & \relv{1.000} & \relv{1.000}    & \tv[n]{91.96}{00.42} & \tv[n]{0.3505}{0.0214} & \tv[n]{0.1366}{0.0074} & \tv[n]{0.0674}{0.0037} & \tv[n]{$<$1}{}      \\

            \gls{ed} (\gls{de}-5)            & \relv{1.000} & \relv{1.000}    & \tv[n]{91.86}{00.21} & \tv[n]{0.3577}{0.0083} & \tv[n]{0.1391}{0.0031} & \tv[n]{0.0683}{0.0017} & \tv[n]{$<$1}{}      \\

            \gls{end2} (\gls{de}-3)        & \relv{1.000} & \relv{1.000}    & \tv[n]{91.99}{00.14} & \tv[n]{0.3405}{0.0079} & \tv[n]{0.1358}{0.0026} & \tv[n]{0.0690}{0.0013} & \tv[n]{$<$1}{}      \\
            
            \gls{end2} (\gls{de}-5)        & \relv{1.000} & \relv{1.000}    & \tv[n]{92.11}{00.23} & \tv[n]{0.3313}{0.0057} & \tv[n]{0.1336}{0.0029} & \tv[n]{0.0645}{0.0014} & \tv[n]{1.077}{}      \\

            \gls{de}-2            & \relv{2.000} & \relv{2.000}    & \tv[n]{92.72}{00.13} & \tv[n]{0.2489}{0.0031} & \tv[n]{0.1125}{0.0012} & \tv[b]{0.0484}{0.0010} & \tv[n]{2.000}{}      \\
            
            \gls{de}-3            & \relv{3.000} & \relv{3.000}    & \tv[b]{93.06}{00.14} & \tv[b]{0.2252}{0.0024} & \tv[b]{0.1038}{0.0008} & \tv[b]{0.0469}{0.0012} & \tv[b]{3.000}{}      \\

            \gls{de}-5            & \relv{5.000} & \relv{5.000}    & \tv[b]{93.61}{00.11} & \tv[b]{0.2005}{0.0015} & \tv[b]{0.0951}{0.0004} & \tv[b]{0.0466}{0.0009} & \tv[b]{5.000}{}      \\
            \bottomrule
           \\
           \textbf{CIFAR-100} \\
            \toprule
            Model                 & \dmet{FLOPs} & \dmet{\#Params} & \umet{ACC}          & \dmet{NLL}           & \dmet{BS}           & \dmet{ECE}            & \umet{DEE}           \\
            
            \midrule

            ResNet (\gls{de}-1)   & \relv{1.000} & \relv{1.000}    & \tv[n]{72.29}{00.36} & \tv[n]{1.1506}{0.0100} & \tv[n]{0.4001}{0.0044} & \tv[n]{0.1526}{0.0005} & \tv[n]{1.000}{}      \\
            
            \metricrule

            \quad +2 \gls{bn}\textsubscript{medium} (\gls{de}-3)  & \relv{1.419} & \relv{1.320}    & \tv[n]{74.97}{00.05} & \tv[n]{1.0360}{0.0022} & \tv[n]{0.3500}{0.0007} & \tv[b]{0.1228}{0.0210} & \tv[n]{1.642}{}      \\

            \quad +1 \gls{dbn} (\gls{de}-3)  & \relv{1.166} & \relv{1.213}    & \tv[b]{76.02}{00.11} & \tv[b]{0.9434}{0.0051} & \tv[b]{0.3438}{0.0009} & \tv[n]{0.1352}{0.0020} & \tv[b]{2.461}{}      \\

            \quad +2 \gls{dbn} (\gls{de}-5) & \relv{1.332} & \relv{1.426}    & \tv[b]{76.82}{00.22} & \tv[b]{0.8998}{0.0046} & \tv[b]{0.3305}{0.0018} & \tv[b]{0.1269}{0.0013} & \tv[b]{3.297}{}      \\

            \gls{ed} (\gls{de}-3)            & \relv{1.000} & \relv{1.000}    & \tv[n]{74.18}{00.22} & \tv[n]{1.2375}{0.0125} & \tv[n]{0.4095}{0.0019} & \tv[n]{0.1819}{0.0013} & \tv[n]{$<$1}{}      \\

            \gls{ed} (\gls{de}-5)            & \relv{1.000} & \relv{1.000}    & \tv[n]{74.00}{00.29} & \tv[n]{1.2428}{0.0174} & \tv[n]{0.4120}{0.0041} & \tv[n]{0.1840}{0.0022} & \tv[n]{$<$1}{}      \\

            \gls{end2} (\gls{de}-3)        & \relv{1.000} & \relv{1.000}    & \tv[n]{73.35}{00.22} & \tv[n]{1.3572}{0.0079} & \tv[n]{0.4350}{0.0013} & \tv[n]{0.1973}{0.0006} & \tv[n]{$<$1}{}      \\
            
            \gls{end2} (\gls{de}-5)        & \relv{1.000} & \relv{1.000}    & \tv[n]{73.22}{00.33} & \tv[n]{1.3597}{0.0156} & \tv[n]{0.4370}{0.0052} & \tv[n]{0.1980}{0.0030} & \tv[n]{$<$1}{}      \\

            \gls{de}-2            & \relv{2.000} & \relv{2.000}    & \tv[n]{74.98}{00.42} & \tv[n]{0.9721}{0.0099} & \tv[n]{0.3505}{0.0034} & \tv[b]{0.1259}{0.0021} & \tv[n]{2.000}{}      \\
            
            \gls{de}-3            & \relv{3.000} & \relv{3.000}    & \tv[b]{76.04}{00.13} & \tv[b]{0.9098}{0.0019} & \tv[b]{0.3342}{0.0007} & \tv[b]{0.1233}{0.0020} & \tv[b]{3.000}{}      \\

            \gls{de}-5            & \relv{5.000} & \relv{5.000}    & \tv[b]{77.03}{00.08} & \tv[b]{0.8606}{0.0036} & \tv[b]{0.3216}{0.0013} & \tv[b]{0.1234}{0.0019} & \tv[b]{5.000}{}      \\
            
            \bottomrule
            \\
            \textbf{TinyImageNet}\\
            \toprule
            Model                 & \dmet{FLOPs} & \dmet{\#Params} & \umet{ACC}          & \dmet{NLL}           & \dmet{BS}           & \dmet{ECE}            & \umet{DEE}           \\
            
            \midrule
            
            ResNet (\gls{de}-1)   & \relv{1.000} & \relv{1.000}    & \tv[n]{59.28}{00.46} & \tv[n]{1.8356}{0.0186} & \tv[n]{0.5426}{0.0052} & \tv[n]{0.2024}{0.0025} & \tv[n]{1.000}{}      \\
            
            \metricrule

            \quad +2 \gls{bn}\textsubscript{medium} (\gls{de}-3)  & \relv{1.359} & \relv{1.412}    & \tv[n]{64.07}{0.16} & \tv[b]{1.5066}{0.0021} & \tv[n]{0.4745}{0.0007} & \tv[b]{0.1267}{0.0137} & \tv[b]{4.102}{}      \\

            \quad +1 \gls{dbn} (\gls{de}-3)  & \relv{1.209} & \relv{1.149}    & \tv[b]{64.60}{00.06} & \tv[n]{1.5299}{0.0022} & \tv[b]{0.4718}{0.0013} & \tv[n]{0.1836}{0.0021} & \tv[n]{3.444}{}      \\

            \quad +2 \gls{dbn} (\gls{de}-5) & \relv{1.418} & \relv{1.298}    & \tv[b]{65.34}{00.10} & \tv[b]{1.4890}{0.0045} & \tv[b]{0.4616}{0.0016} & \tv[n]{0.1803}{0.0019} & \tv[b]{4.701}{}      \\

            \gls{ed} (\gls{de}-3)            & \relv{1.000} & \relv{1.000}    & \tv[n]{60.81}{00.26} & \tv[n]{1.8312}{0.0129} & \tv[n]{0.5443}{0.0052} & \tv[n]{0.2126}{0.0021} & \tv[n]{$<$1}{}      \\

            \gls{end2} (\gls{de}-3)        & \relv{1.000} & \relv{1.000}    & \tv[n]{60.71}{00.31} & \tv[n]{1.9991}{0.0165} & \tv[n]{0.5828}{0.0026} & \tv[n]{0.2279}{0.0010} & \tv[n]{$<$1}{}      \\

            \gls{de}-2            & \relv{2.000} & \relv{2.000}    & \tv[n]{62.47}{00.35} & \tv[n]{1.6264}{0.0110} & \tv[n]{0.4942}{0.0033} & \tv[n]{0.1834}{0.0013} & \tv[n]{2.000}{}      \\
            
            \gls{de}-3            & \relv{3.000} & \relv{3.000}    & \tv[n]{63.78}{00.23} & \tv[n]{1.5538}{0.0055} & \tv[n]{0.4788}{0.0015} & \tv[n]{0.1836}{0.0021} & \tv[n]{3.000}{}      \\
            
            \bottomrule

        \end{tabular}
    }
    \label{tab:classification}
\end{table*}

\begin{figure}[t]
\minipage{0.55\textwidth}
    \centering
    \includegraphics[width=1\textwidth]{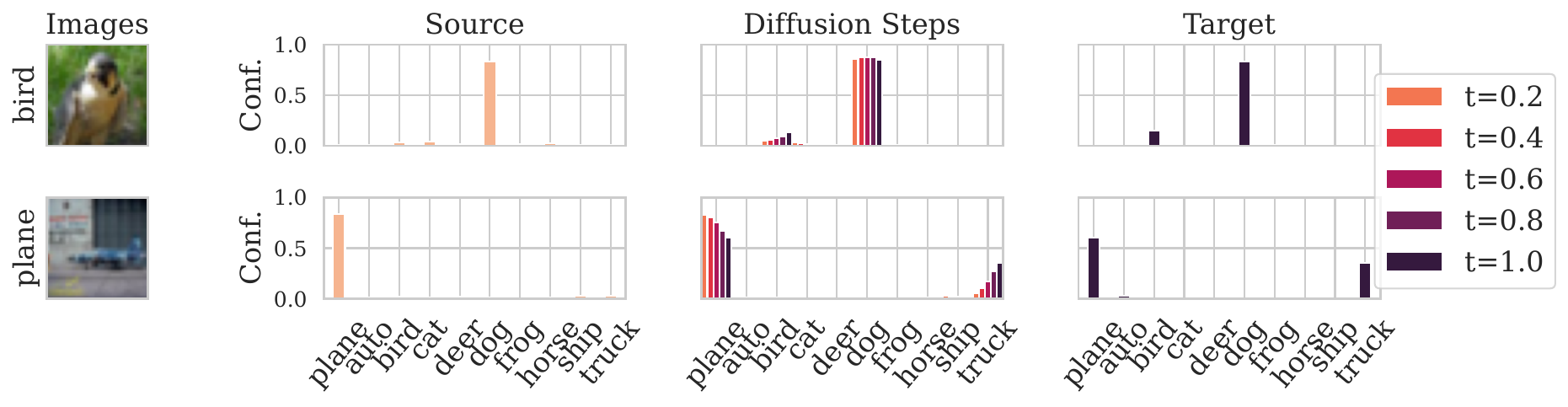}
    \vspace{-7mm}
    \caption{Confidences from the source model (first column), from the ensemble model (third column), and from the diffusion bridge (middle column) in the CIFAR-10 dataset. The middle column illustrates a transition of the diffusion process.}
    \label{fig:confidence}
\endminipage\hfill
\minipage{0.4\textwidth}%
    \centering
    \includegraphics[width=\textwidth]{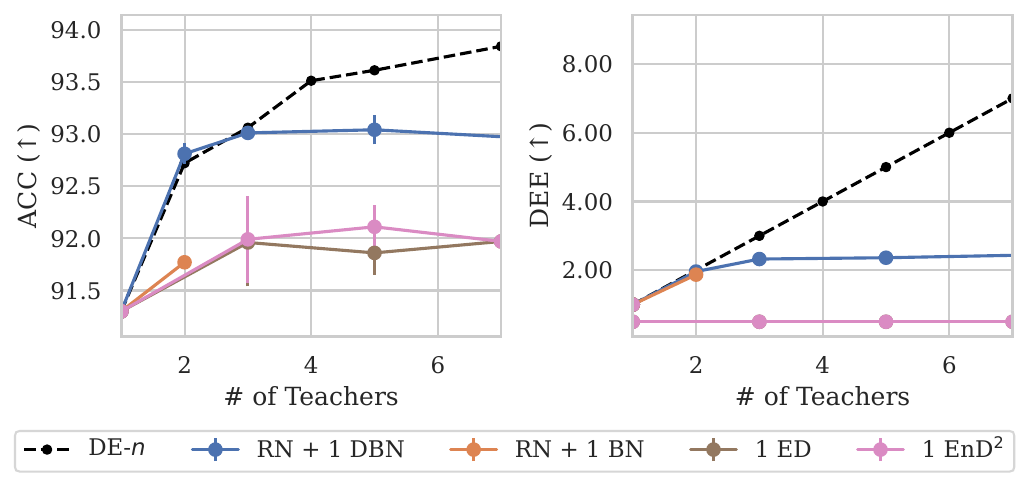}
    \vspace{-7mm}
    \caption{The number of teachers that a single model can distill in terms of ACC (left) and DEE (right). DEEs less than 1 are set to 0.5.}
    \label{fig:capacity}
\endminipage
\end{figure}

The ACC, NLL, BS, ECE, and DEE comparisons with the baselines with respect to \gls{flops} and \#Params on CIFAR-10, CIFAR-100 and TinyImageNet are shown in~\cref{tab:classification}. We assume the situation where both \gls{bn} and \gls{dbn} can utilize only a single source model to make the problem difficult. As we can see in the results of CIFAR-10 and CIFAR-100, with only a small increase in the computational costs (\gls{flops} and \#Params), \gls{dbn} achieves almost \gls{de}-3 performance, whereas \gls{bn} struggles to achieve even \gls{de}-2 performance with more computational costs than \gls{dbn}. In TinyImageNet, \gls{dbn} even outperforms \gls{de}-3 with less than a half of computation costs. On the other hand, the two distillation methods, \gls{ed} and \gls{end2}, are competitive with \gls{bn} but they shows poor uncertainty metrics such as NLL, BS, and ECE, and interestingly \gls{dbn} also shows poor ECE scores even with high performance in the other uncertainty metrics. In addition, the actual output results through the diffusion processes are illustrated in~\cref{fig:confidence} and the results for the other images are listed in~\cref{app:sec:transitions}.

\subsection{Performance Loss vs. Computational Efficiency}
\label{main:sec:tradeoff}

\begin{figure}[t]
    \centering
    \includegraphics[width=\textwidth]{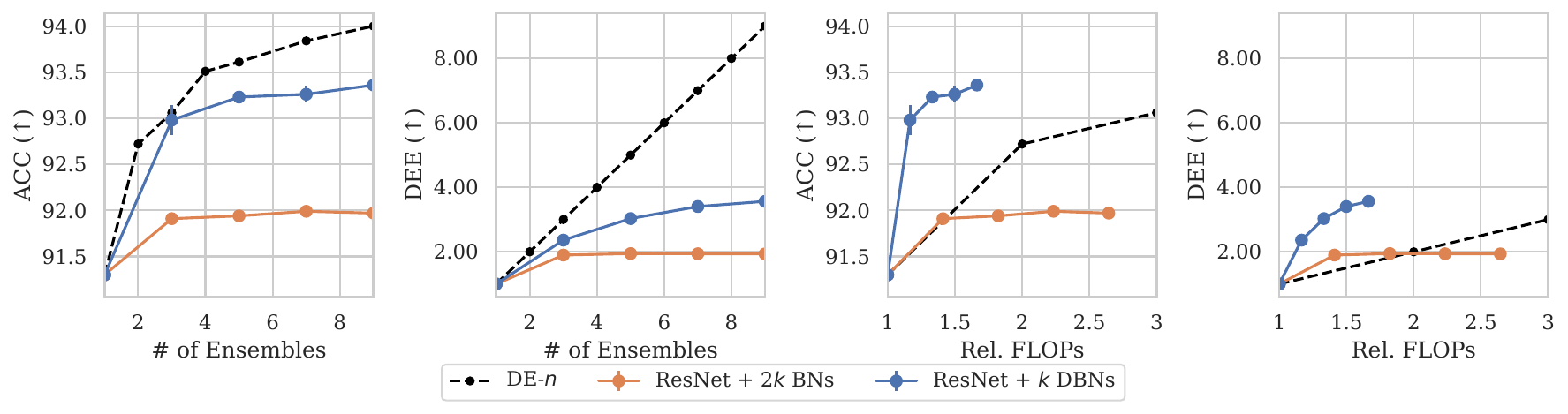}
    \vspace{-3mm}
    \caption{From \textbf{left} (1$^\text{st}$) to \textbf{right} (4$^\text{th}$): Number of target ensembles vs. accuracy (1$^\text{st}$) or DEE (2$^\text{nd}$), and relative FLOPs vs. accuracy (3$^\text{rd}$) or DEE (4$^\text{th}$) in CIFAR-10.}
    \label{fig:dee}
\end{figure}

Furthermore, depending on the number of ensemble models that \gls{dbn} is trained on, the left hand side of \cref{fig:dee} demonstrates how much performance loss occurs compared to the \gls{de} when the number of target ensemble models increase. Conversely, the right hand side of \cref{fig:dee} illustrates how much cost savings DBN can achieve compared to \gls{de}, given the same computational cost (\gls{flops}). The comparison is made from the perspectives of ACC and DEE. In this experiment, a maximum of 9 ensemble models are used and we also compare with our competing model, \gls{bn}. \gls{dbn} trains one diffusion bridge with three ensembles, while \gls{bn} learns a low-loss curve between 2 ensembles for one bridge. For more than four ensembles, \gls{dbn} conducts ensemble inference aggregating two or more diffusion bridges as \gls{bn} does. We use the standard size of \gls{bn} in "ResNet + 2 \glspl{bn}" which has 1.411 relative FLOPs for a single network compared to "ResNet + 1 \glspl{dbn}" which has 1.166 relative FLOPs. As shown in~\cref{fig:dee}, \gls{dbn} (left) achieves significant ensemble gains even when the number of target ensemble increases, whereas \gls{bn} (left) saturates at ACC 92.0\% and DEE 2\%. Moreover, \gls{dbn} (right) shows rapid ensemble inference compared to the target ensemble \gls{de} (right) and even faster than \gls{bn} (right) to get the same ensemble performance.

\subsection{Ensemble Capacity of a Single DBN}
\label{main:sec:capacity}
\cref{fig:capacity} demonstrates the ensemble capacity of how much a single~\gls{dbn} model learns and distills knowledge from multiple teacher (ensemble) models in terms of ACC and DEE in the CIFAR-10 dataset.
Existing ensemble distillation models have limited performance in learning likelihood and are saturated in terms of ACC at learning at most 3 number of target ensembles.
On the other hand, ~\gls{bn} assumes that a single model can cover at most two ensemble modes, hence the maximum gain on ACC and DEE is capped at two.
Compared to these competing methods, our~\gls{dbn} method succeeds in learning slightly less than three ensembles only with a single lightweight model and the source model.

\section{Related Work}
\label{main:sec:related-work}

\textit{Fast Ensembling Methods for Neural Network Prediction.}
Many works have suggested to reduce the computational cost of the typical ensemble methods that are multiplied proportional to the number of the ensemble networks.
To provide output diversity with lightweight computation, sharing either weights or latent features of neural networks enabled to compress redundancy throughout the ensemble members and add minor differences between them  \citep{wen2019batchensemble, dusenberry2020efficient, lee2015m, siqueira2018ensemble, antoran2020depth, havasi2020training}.
This can also be interpreted as the knowledge distillation from the ensemble of networks to the single network; \citep{hinton2015distilling}, \citet{malinin2020ensemble, ryabinin2021scaling, penso2022functional} showed that distilling deep ensemble to a single network helps retaining information of the ensemble distribution without heavy computational burden.

\textit{Restoration with Conditional Diffusion Models.}
Our model is on the line of the reconstruction problem from some degraded measurement, if we consider our method as reconstructing the ensemble distribution from the degraded function output from a single model.
This line of work starts with the conditional diffusion model~\citep{saharia2022palette}, that refines the image given some conditional features.
Conditional diffusion models have achieved success in various problems such as time series imputation~\citep{tashiro2021csdi}, deblurring~\citep{whang2022deblurring}, and super-resolution~\citep{saharia2023}.
As a generalized perspective, inverse problems dealt with diffusion~\citep{song2022inverse} aims to restore the underlying clean signal from the noisy measurement.
\citet{wang2022zero} delves into the null space of the image and utilize the valid space that should be recovered, and achieved zero-shot restoration using diffusion models.
\section{Conclusion}
\label{main:sec:conclusion}

We have proposed a novel approach to approximate the performance of Deep Ensembles with reduced computational cost. To achieve this, we constructed Conditional Diffusion Bridge that connects the logit distribution between one of the ensemble models and the target ensemble. We approximated the output distribution of the target ensemble using the features and logits obtained from the source model. The computations involved in this process consist of a single forward pass of the source model and the diffusion process with a lightweight score network. Additionally, during the training process, distilling the multiple diffusion steps into one step accelerates the inference speed while retaining the performance of three deep ensemble models. We evaluated this method on three widely used datasets and achieved superior performance compared to the baselines. Notably, we demonstrated significantly faster computations compared to competitive models such as Bridge Network while achieving similar performance. However, there are still some limitations to consider. First, a single Diffusion Bridge still has limitations in terms of the number of ensembles it can learn. If more ensembles are required, additional diffusion bridges must be used. Secondly, the use of multiple diffusion bridges leads to a proportional training time because diffusion models demand a long training time due to their multiple diffusion steps.


\pagebreak
\newpage
\subsubsection*{Acknowledgments}
This work is partly supported by Institute for Information \& communications Technology Planning \& Evaluation (IITP) grant funded by the Korea government (MSIT) (No.2019-0-00075: Artificial Intelligence Graduate School Program (KAIST), No.2022-0-00713: Meta-learning Applicable to Real-world Problems, No.2022-0-00184: Development and Study of AI Technologies to Inexpensively Conform to Evolving Policy on Ethics), and KAIST-NAVER Hypercreative AI Center.
Our research is supported with Cloud TPUs from Google's TPU Research Cloud (TRC).

\paragraph{Ethics statement.} This paper does not include any ethical issues. This paper presents a fast ensemble inference algorithm of mainly image classifications which does not cause ethical issues.
\paragraph{Reproducibility statement.} We described our experimental details in~\cref{app:sec:experiment} and~\cref{app:sec:metrics} which includes information about datasets, architectures, and hyperparameters used.

\bibliography{iclr2024_conference}

\begin{thebibliography}{39}
\providecommand{\natexlab}[1]{#1}
\providecommand{\url}[1]{\texttt{#1}}
\expandafter\ifx\csname urlstyle\endcsname\relax
  \providecommand{\doi}[1]{doi: #1}\else
  \providecommand{\doi}{doi: \begingroup \urlstyle{rm}\Url}\fi

\bibitem[Antor{\'a}n et~al.(2020)Antor{\'a}n, Allingham, and
  Hern{\'a}ndez-Lobato]{antoran2020depth}
Javier Antor{\'a}n, James~Urquhart Allingham, and Jos{\'e}~Miguel
  Hern{\'a}ndez-Lobato.
\newblock Depth uncertainty in neural networks.
\newblock In \emph{Advances in Neural Information Processing Systems 33
  (NeurIPS 2020)}, 2020.

\bibitem[Ashukha et~al.(2020)Ashukha, Lyzhov, Molchanov, and
  Vetrov]{ashukha2020pitfalls}
Arsenii Ashukha, Alexander Lyzhov, Dmitry Molchanov, and Dmitry~P. Vetrov.
\newblock Pitfalls of in-domain uncertainty estimation and ensembling in deep
  learning.
\newblock In \emph{International Conference on Learning Representations
  (ICLR)}, 2020.

\bibitem[Bortoli et~al.(2021)Bortoli, Thornton, Heng, and
  Doucet]{bortoli2021diffusion}
Valentin~De Bortoli, James Thornton, Jeremy Heng, and Arnaud Doucet.
\newblock Diffusion schr{\"{o}}dinger bridge with applications to score-based
  generative modeling.
\newblock In \emph{Advances in Neural Information Processing Systems 34
  (NeurIPS 2021)}, 2021.

\bibitem[Breiman(1996)]{breiman96}
Leo Breiman.
\newblock Bagging predictors.
\newblock \emph{Machine Learning}, 1996.

\bibitem[{Brier}(1950)]{brier1950verification}
Glenn~W. {Brier}.
\newblock {Verification of Forecasts Expressed in Terms of Probability}.
\newblock \emph{Monthly Weather Review}, 1950.

\bibitem[Chen et~al.(2022)Chen, Liu, and Theodorou]{chen2023likelihood}
Tianrong Chen, Guan{-}Horng Liu, and Evangelos~A. Theodorou.
\newblock Likelihood training of schr{\"{o}}dinger bridge using
  forward-backward sdes theory.
\newblock In \emph{International Conference on Learning Representations
  (ICLR)}, 2022.

\bibitem[Chollet(2017)]{chollet2017xception}
Fran{\c{c}}ois Chollet.
\newblock Xception: Deep learning with depthwise separable convolutions.
\newblock In \emph{Proceedings of the IEEE Conference on Computer Vision and
  Pattern Recognition (CVPR)}, 2017.

\bibitem[Dusenberry et~al.(2020)Dusenberry, Jerfel, Wen, Ma, Snoek, Heller,
  Lakshminarayanan, and Tran]{dusenberry2020efficient}
Michael Dusenberry, Ghassen Jerfel, Yeming Wen, Yian Ma, Jasper Snoek,
  Katherine Heller, Balaji Lakshminarayanan, and Dustin Tran.
\newblock Efficient and scalable bayesian neural nets with rank-1 factors.
\newblock In \emph{Proceedings of The 37th International Conference on Machine
  Learning (ICML 2020)}, 2020.

\bibitem[Garipov et~al.(2018)Garipov, Izmailov, Podoprikhin, Vetrov, and
  Wilson]{garipov2018loss}
Timur Garipov, Pavel Izmailov, Dmitrii Podoprikhin, Dmitry Vetrov, and
  Andrew~Gordon Wilson.
\newblock Loss surfaces, mode connectivity, and fast ensembling of {DNNs}.
\newblock In \emph{Advances in Neural Information Processing Systems 31
  (NeurIPS 2018)}, 2018.

\bibitem[Guo et~al.(2017)Guo, Pleiss, Sun, and Weinberger]{guo2017on}
Chuan Guo, Geoff Pleiss, Yu~Sun, and Kilian~Q. Weinberger.
\newblock On calibration of modern neural networks.
\newblock In \emph{Proceedings of The 34th International Conference on Machine
  Learning (ICML 2017)}, 2017.

\bibitem[Havasi et~al.(2021)Havasi, Jenatton, Fort, Liu, Snoek,
  Lakshminarayanan, Dai, and Tran]{havasi2020training}
Marton Havasi, Rodolphe Jenatton, Stanislav Fort, Jeremiah~Zhe Liu, Jasper
  Snoek, Balaji Lakshminarayanan, Andrew~M Dai, and Dustin Tran.
\newblock Training independent subnetworks for robust prediction.
\newblock In \emph{International Conference on Learning Representations
  (ICLR)}, 2021.

\bibitem[He et~al.(2016)He, Zhang, Ren, and Sun]{he2016deep}
Kaiming He, Xiangyu Zhang, Shaoqing Ren, and Jian Sun.
\newblock Deep residual learning for image recognition.
\newblock In \emph{Proceedings of the IEEE Conference on Computer Vision and
  Pattern Recognition (CVPR)}, 2016.

\bibitem[Hendrycks \& Dietterich(2019)Hendrycks and
  Dietterich]{hendrycks2019benchmarking}
Dan Hendrycks and Thomas~G. Dietterich.
\newblock Benchmarking neural network robustness to common corruptions and
  perturbations.
\newblock In \emph{International Conference on Learning Representations
  (ICLR)}, 2019.

\bibitem[Hinton et~al.(2015)Hinton, Vinyals, and Dean]{hinton2015distilling}
Geoffrey~E. Hinton, Oriol Vinyals, and Jeffrey Dean.
\newblock Distilling the knowledge in a neural network.
\newblock In \emph{Advances in Neural Information Processing Systems 27 (NIPS
  2014)}, 2015.

\bibitem[Ho et~al.(2020)Ho, Jain, and Abbeel]{ho2020ddpm}
Jonathan Ho, Ajay Jain, and Pieter Abbeel.
\newblock Denoising diffusion probabilistic models.
\newblock In \emph{Advances in Neural Information Processing Systems 33
  (NeurIPS 2020)}, 2020.

\bibitem[Kingma \& Ba(2015)Kingma and Ba]{kingma2015adam}
Diederik~P. Kingma and Jimmy Ba.
\newblock Adam: {A} method for stochastic optimization.
\newblock In \emph{International Conference on Learning Representations
  (ICLR)}, 2015.

\bibitem[Krizhevsky et~al.(2009)Krizhevsky, Hinton,
  et~al.]{Krizhevsky09learningmultiple}
Alex Krizhevsky, Geoffrey Hinton, et~al.
\newblock Learning multiple layers of features from tiny images, 2009.

\bibitem[Lakshminarayanan et~al.(2017)Lakshminarayanan, Pritzel, and
  Blundell]{lakshminarayanan2017simple}
Balaji Lakshminarayanan, Alexander Pritzel, and Charles Blundell.
\newblock Simple and scalable predictive uncertainty estimation using deep
  ensembles.
\newblock In \emph{Advances in Neural Information Processing Systems 30 (NIPS
  2017)}, 2017.

\bibitem[Lee et~al.(2015)Lee, Purushwalkam, Cogswell, Crandall, and
  Batra]{lee2015m}
Stefan Lee, Senthil Purushwalkam, Michael Cogswell, David Crandall, and Dhruv
  Batra.
\newblock Why m heads are better than one: Training a diverse ensemble of deep
  networks.
\newblock \emph{arXiv:1511.06314}, 2015.

\bibitem[Li et~al.(2017)Li, Karpathy, and Johnson]{tinyimagenet}
Fei-Fei Li, Andrej Karpathy, and Justin Johnson.
\newblock {Tiny ImageNet}.
\newblock \url{https://www.kaggle.com/c/tiny-imagenet}, 2017.
\newblock [Online; accessed 19-May-2022].

\bibitem[Liu et~al.(2023)Liu, Vahdat, Huang, Theodorou, Nie, and
  Anandkumar]{liu23i2sb}
Guan{-}Horng Liu, Arash Vahdat, De{-}An Huang, Evangelos~A. Theodorou, Weili
  Nie, and Anima Anandkumar.
\newblock I\({}^{\mbox{2}}\)sb: Image-to-image schr{\"{o}}dinger bridge.
\newblock In \emph{Proceedings of The 40th International Conference on Machine
  Learning (ICML 2023)}, 2023.

\bibitem[Malinin et~al.(2020)Malinin, Mlodozeniec, and
  Gales]{malinin2020ensemble}
Andrey Malinin, Bruno Mlodozeniec, and Mark J.~F. Gales.
\newblock Ensemble distribution distillation.
\newblock In \emph{International Conference on Learning Representations
  (ICLR)}, 2020.

\bibitem[Nam et~al.(2021)Nam, Yoon, Lee, and Lee]{nam2021diversity}
Giung Nam, Jongmin Yoon, Yoonho Lee, and Juho Lee.
\newblock Diversity matters when learning from ensembles.
\newblock In \emph{Advances in Neural Information Processing Systems 34
  (NeurIPS 2021)}, 2021.

\bibitem[Penso et~al.(2022)Penso, Achituve, and Fetaya]{penso2022functional}
Coby Penso, Idan Achituve, and Ethan Fetaya.
\newblock Functional ensemble distillation.
\newblock In \emph{Advances in Neural Information Processing Systems 35
  (NeurIPS 2022)}, 2022.

\bibitem[Ryabinin et~al.(2021)Ryabinin, Malinin, and
  Gales]{ryabinin2021scaling}
Max Ryabinin, Andrey Malinin, and Mark Gales.
\newblock Scaling ensemble distribution distillation to many classes with proxy
  targets.
\newblock In \emph{Advances in Neural Information Processing Systems 34
  (NeurIPS 2021)}, 2021.

\bibitem[Saharia et~al.(2022)Saharia, Chan, Chang, Lee, Ho, Salimans, Fleet,
  and Norouzi]{saharia2022palette}
Chitwan Saharia, William Chan, Huiwen Chang, Chris~A. Lee, Jonathan Ho, Tim
  Salimans, David~J. Fleet, and Mohammad Norouzi.
\newblock Palette: Image-to-image diffusion models.
\newblock In \emph{{SIGGRAPH} (Conference Paper Track)}, 2022.

\bibitem[Saharia et~al.(2023)Saharia, Ho, Chan, Salimans, Fleet, and
  Norouzi]{saharia2023}
Chitwan Saharia, Jonathan Ho, William Chan, Tim Salimans, David~J. Fleet, and
  Mohammad Norouzi.
\newblock Image super-resolution via iterative refinement.
\newblock \emph{{IEEE} Trans. Pattern Anal. Mach. Intell.}, 2023.

\bibitem[Salimans \& Ho(2022)Salimans and Ho]{salimans2022progressive}
Tim Salimans and Jonathan Ho.
\newblock Progressive distillation for fast sampling of diffusion models.
\newblock In \emph{International Conference on Learning Representations
  (ICLR)}, 2022.

\bibitem[Sandler et~al.(2018)Sandler, Howard, Zhu, Zhmoginov, and
  Chen]{sandler2018mobilenetv2}
Mark Sandler, Andrew~G. Howard, Menglong Zhu, Andrey Zhmoginov, and
  Liang{-}Chieh Chen.
\newblock Mobilenetv2: Inverted residuals and linear bottlenecks.
\newblock In \emph{Proceedings of the IEEE Conference on Computer Vision and
  Pattern Recognition (CVPR)}, 2018.

\bibitem[Schrödinger(1932)]{sch1932}
E.~Schrödinger.
\newblock Sur la théorie relativiste de l'électron et l'interprétation de la
  mécanique quantique.
\newblock \emph{Annales de l'institut Henri Poincaré}, 1932.

\bibitem[Singh \& Krishnan(2020)Singh and Krishnan]{singh2020filter}
Saurabh Singh and Shankar Krishnan.
\newblock Filter response normalization layer: Eliminating batch dependence in
  the training of deep neural networks.
\newblock In \emph{Proceedings of the IEEE Conference on Computer Vision and
  Pattern Recognition (CVPR)}, 2020.

\bibitem[Siqueira et~al.(2018)Siqueira, Barros, Magg, and
  Wermter]{siqueira2018ensemble}
Henrique Siqueira, Pablo Barros, Sven Magg, and Stefan Wermter.
\newblock An ensemble with shared representations based on convolutional
  networks for continually learning facial expressions.
\newblock In \emph{2018 IEEE/RSJ International Conference on Intelligent Robots
  and Systems (IROS)}, 2018.

\bibitem[Song et~al.(2022)Song, Shen, Xing, and Ermon]{song2022inverse}
Yang Song, Liyue Shen, Lei Xing, and Stefano Ermon.
\newblock Solving inverse problems in medical imaging with score-based
  generative models.
\newblock In \emph{International Conference on Learning Representations
  (ICLR)}, 2022.

\bibitem[Tashiro et~al.(2021)Tashiro, Song, Song, and Ermon]{tashiro2021csdi}
Yusuke Tashiro, Jiaming Song, Yang Song, and Stefano Ermon.
\newblock {CSDI:} conditional score-based diffusion models for probabilistic
  time series imputation.
\newblock In \emph{Advances in Neural Information Processing Systems 34
  (NeurIPS 2021)}, 2021.

\bibitem[Tran et~al.(2021)Tran, Veeling, Roth, Swiatkowski, Dillon, Snoek,
  Mandt, Salimans, Nowozin, and Jenatton]{tran2021hydra}
Linh Tran, Bastiaan~S. Veeling, Kevin Roth, Jakub Swiatkowski, Joshua~V.
  Dillon, Jasper Snoek, Stephan Mandt, Tim Salimans, Sebastian Nowozin, and
  Rodolphe Jenatton.
\newblock Hydra: Preserving ensemble diversity for model distillation, 2021.

\bibitem[Wang et~al.(2023)Wang, Yu, and Zhang]{wang2022zero}
Yinhuai Wang, Jiwen Yu, and Jian Zhang.
\newblock Zero-shot image restoration using denoising diffusion null-space
  model.
\newblock In \emph{International Conference on Learning Representations
  (ICLR)}, 2023.

\bibitem[Wen et~al.(2019)Wen, Tran, and Ba]{wen2019batchensemble}
Yeming Wen, Dustin Tran, and Jimmy Ba.
\newblock Batchensemble: an alternative approach to efficient ensemble and
  lifelong learning.
\newblock In \emph{International Conference on Learning Representations
  (ICLR)}, 2019.

\bibitem[Whang et~al.(2022)Whang, Delbracio, Talebi, Saharia, Dimakis, and
  Milanfar]{whang2022deblurring}
Jay Whang, Mauricio Delbracio, Hossein Talebi, Chitwan Saharia, Alexandros~G.
  Dimakis, and Peyman Milanfar.
\newblock Deblurring via stochastic refinement.
\newblock In \emph{Proceedings of the IEEE Conference on Computer Vision and
  Pattern Recognition (CVPR)}, 2022.

\bibitem[Yun et~al.(2023)Yun, Lee, Nam, and Lee]{yun2023traversing}
EungGu Yun, Hyungi Lee, Giung Nam, and Juho Lee.
\newblock Traversing between modes in function space for fast ensembling.
\newblock In \emph{Proceedings of The 40th International Conference on Machine
  Learning (ICML 2023)}, 2023.

\end{thebibliography}
\bibliographystyle{iclr2024_conference}

\newpage

\appendix
\section{Experimental Details}
\label{app:sec:experiment}

We describe the overall details of our experiment below.
\subsection{Datasets}
We employ CIFAR-10/100 \citep{Krizhevsky09learningmultiple}, and TinyImageNet \citep{tinyimagenet} datasets for our study. Our data augmentation strategy involves randomly cropping images by 32 pixels with an additional 4-pixel padding, as well as applying random horizontal flipping. Furthermore, we normalize input images by subtracting per-channel means and dividing them by per-channel standard deviations.

\subsection{Target Model Architectures}
In our investigation, we implement comparable ResNet block configurations. The overall network architectures are consistent with one of our baseline, \gls{bn} \citep{yun2023traversing}, to compare in a reliable condition with a minor difference in TinyImageNet and ImageNet.

\paragraph{Classifier Architectures.}

\paragraph{CIFAR-10.} ResNet-$32\times 2$, characterized by 15 basic blocks distributed as (5, 5, 5) and a total of 32 layers with the Filter Response normalization (FRN)~\citep{singh2020filter} and the Swish activation layer. This model incorporates a widen factor of 2 and operates with in-planes set at 16.

\paragraph{CIFAR-100.} ResNet-$32\times 4$, which closely resembles the CIFAR-10 network with a widen factor of 4 with FRN and the Swish activation.

\paragraph{TinyImageNet.} ResNet-34, which encompasses 16 basic blocks organized as (3, 4, 6, 3) and 34 layers in total with FRN and Swish. The in-planes parameter for this model is established at 64. The only difference with the target model of \citet{yun2023traversing} is that we use bias in the convolutional layers but they don't.

\paragraph{ImageNet64.} ResNet-34, which encompasses 16 basic blocks organized as (3, 4, 6, 3) and 34 layers in total with FRN and Swish. The in-planes parameter for this model is established at 64 as in TinyImageNet. However, we don't use bias in every convolutional layers in this task.

\begin{figure*}[h]
    \centering
    \includegraphics[width=0.65\textwidth]{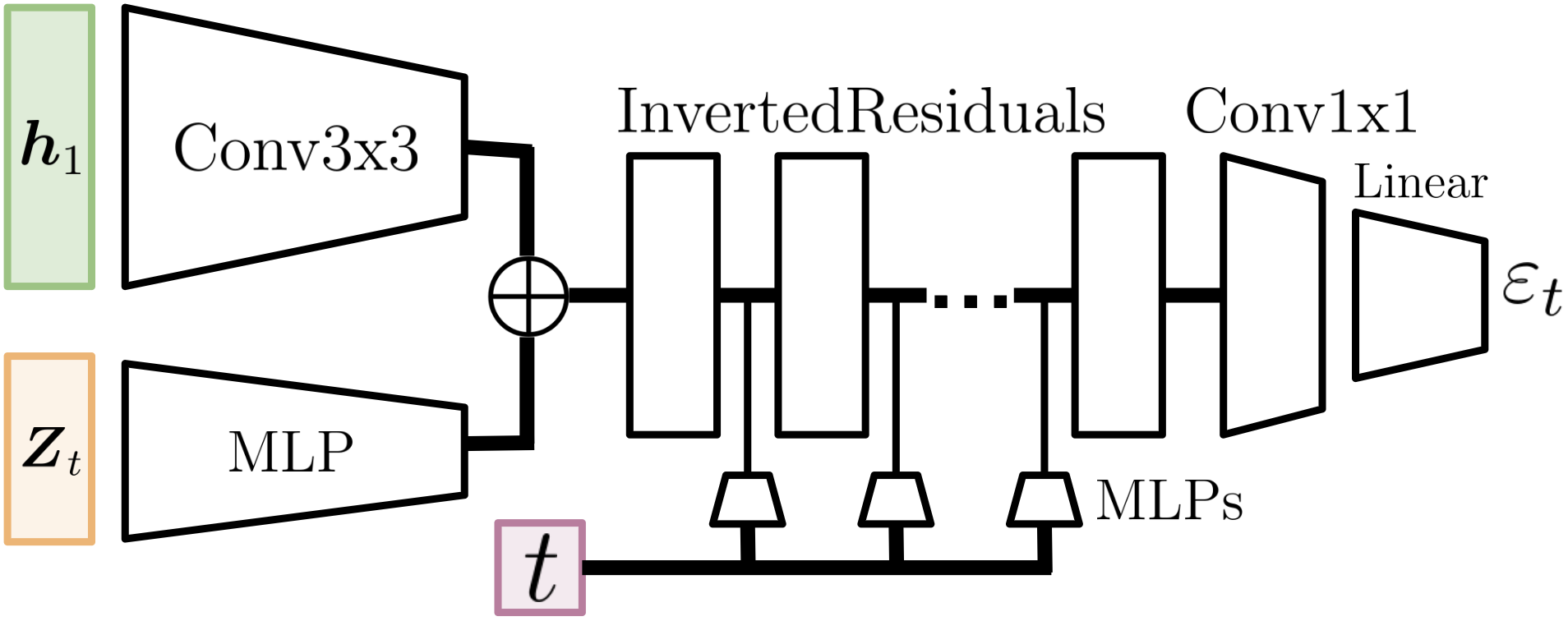}
    \vspace{-3mm}
    \caption{Score network architecture.}
    \label{fig:scorenet}
\end{figure*}

\paragraph{Score Network Architectures.} The score networks consist of three embedding networks for $\bsh_1$, $\bsZ_t$, and $t$, some inverted residual blocks \citep{sandler2018mobilenetv2}, and the output layer. The embedding network for $\bsZ_t$ consists of a LayerNorm layer followed by the average pooling layer, and returns the embedding output by a forward computation to a lightweight MLP (width $2\to 2\to 1$ with the ReLU6 activation layer), and concatenate the embedding outputs to the embeddings of $\bsh_1$. The embedding network for $t$ firstly use the sinusoidal time-step embeddings with the maximum period 10,000 and then feed-forward to a light weight MLP (width $d/4\rightarrow d/2\rightarrow e$ with the Swish activation, where $d$ is the dimension of the logits and $e$ is the input channels of each inverted residual blocks.) The specifications of the inverted residuals and the size of the output layers in the score networks vary across the tasks as in~\cref{tab:invertedresiduals}.

\begin{table}[ht]
\newcommand{\metricrule}{\cmidrule(lr){1-5}\cmidrule(lr){6-10}\cmidrule(lr){11-15}\cmidrule(lr){16-20}}
    \renewcommand{\arraystretch}{1.18}
    \centering
    \caption{
        Specification of the inverted residual blocks and output layers in the score network across the tasks. $t$ is the expansion factor~\citep{sandler2018mobilenetv2}. $c$ is the output channels. $n$ is the number of layers. $s$ is the stride applied in the first layer of each inverted residual.
    }
    \vspace{0.5mm}
    \resizebox{\textwidth}{!}{
        \begin{tabular}{cccccccccccccccccccc}
            \toprule
            \multicolumn{5}{c}{CIFAR10} & \multicolumn{5}{c}{CIFAR100} & \multicolumn{5}{c}{TinyImageNet} & \multicolumn{5}{c}{ImageNet64}
            \\
            \metricrule
            Operator & $t$ & $c$ &$n$ &$s$ & Operator & $t$ & $c$ &$n$ &$s$ & Operator & $t$ & $c$ &$n$ &$s$ & Operator & $t$ & $c$ &$n$ &$s$ \\
            \metricrule
            InvertedResidual & 1 & 32 & 1 & 1 & InvertedResidual & 1 & 64 & 1 & 1 & InvertedResidual & 1 & 64 & 1 & 1 & InvertedResidual & 6 & 64 & 3 & 2 \\
            InvertedResidual & 3 & 48 & 2 & 2 & InvertedResidual & 3 & 96 & 2 & 2 & InvertedResidual & 4 & 96 & 2 & 2 & InvertedResidual & 6 & 96 & 3 & 2 \\
            InvertedResidual & 3 & 64 & 3 & 2 & InvertedResidual & 3 & 128 & 3 & 2 & InvertedResidual & 4 & 128 & 3 & 2 & InvertedResidual & 6 & 160 & 3 & 1 \\
            InvertedResidual & 3 & 96 & 2 & 1 & InvertedResidual & 3 & 192 & 2 & 1 & InvertedResidual & 4 & 192 & 4 & 1 & InvertedResidual & 6 & 320 & 3 & 2 \\
            InvertedResidual & 3 & 128 & 2 & 1 & InvertedResidual & 3 & 256 & 2 & 1 & InvertedResidual & 4 & 256 & 3 & 2 & InvertedResidual & 6 & 640 & 1 & 1 \\
            Conv 1x1 & - & 128 & 1 & 1 & Conv 1x1 & - & 256 & 1 & 1 & Conv 1x1 & - & 256 & 1 & 1 & Conv 1x1 & - & 1280 & 1 & 1 \\
            AvgPool & - & - & 1 & - & AvgPool & - & - & 1 & - & AvgPool & - & - & 1 & - & AvgPool & - & - & 1 & - \\
            Linear & - & 10 & 1 & - & Linear & - & 100 & 1 & - & Linear & - & 200 & 1 & - & Linear & - & 1000 & 1 & - 
            \hspace{3mm}
            \\
            \bottomrule
        \end{tabular}
    }
    \label{tab:invertedresiduals}
\end{table}

\subsection{Metrics}
\label{app:sec:metrics}
We introduce the metrics used in our experiments.
\begin{itemize}
\item
Accuracy
    \[
    \bbE_{(\bsx,y)} \Big[ I[
        y = \argmax_k \bsp^{(k)}(\bsx)
    ] \Big],
    \] where $I$ is the indicator function.
\item 
Negative log-likelihood (NLL)
    \[
    \bbE_{(\bsx,y)} \Big[
        -\log{\bsp^{(y)}(\bsx)}
    \Big].
    \]
\item 
Brier score (BS)
    \[
    \bbE_{(\bsx,y)} \bigg[
        \Big\lVert \bsp(\bsx) - \bsy \Big\rVert_2^2
    \bigg],
    \]where $\bsy$ is a one-hot-encoded label $y$.
\item 
Expected calibration error (ECE)
    \[
    \operatorname{ECE}(\calD, N_{\text{bin}}) = \sum_{b=1}^{N_{\text{bin}}} \frac{
        n_b |\delta_b|
    }{
        n_1 + \cdots + n_{N_{\text{bin}}}
    },
    \] where $N_{\text{bin}}$ is the quantity of bins, $n_b$ is the number of instances within the $b^\mathrm{th}$ bin, and $\delta_b$ is the calibration discrepancy associated with the $b^\mathrm{th}$ bin. To elaborate, the $b^\mathrm{th}$ bin encompasses predictions characterized by the highest confidence levels falling within the interval $[(b-1)/K, b/K)$, and the calibration error is defined as the disparity between accuracy and the mean confidence values. We maintains $N_{\text{bin}}=15$ throughout the paper.
\item 
Deep ensemble equivalent score (DEE)

First introduced in~\citet{ashukha2020pitfalls}, this metric assumes that the NLL of the DE-$k$ model decreases monotonely by $k$, and obtain how equivalent the objective model to how many ensemble of the baseline model, as
\[
\text{DEE}(f)
&=
s + \frac{\text{NLL}(f) - \text{NLL}(f_s)}{\text{NLL}(f_{s+1}) - \text{NLL} (f_s)}, \\
s 
&=
\argmax_i \{i\in\bbN: \text{NLL}(f_i) \geq \text{NLL}(f) \}
\]
where $f_i$ is the DE-$i$ model.
In our paper, DEE is linearly extrapolated below 1 if $\text{NLL}(f) > \text{NLL}(f_1)$.
\end{itemize}
\section{Hyperparameter settings}
\label{app:sec:hp}
We list common hyperparameters for training \glspl{dbn} in every dataset as follows:
\paragraph{Optimizer.}
For training our~\gls{dbn} model, we use the ADAM~\citep{kingma2015adam} optimizer with zero weight decay and apply consine-decay as a learning-rate scheduling.
For training the teacher ensemble model, we followed the BN paper by using the SGD optimizer with weight decay, as followed in~\cref{tab:hp_ensemble}.
\paragraph{Regularization.}
We use ~\gls{ema} with 0.99995 decay factor and the mixup augmentation with $\alpha=0.4$ except for ImageNet, following \citep{yun2023traversing}.
\paragraph{Diffusion Model.}
Our~\glspl{dbn} follow the training policy of discrete-time conditional diffusion model with uniform timesteps.
After training the baseline~\glspl{dbn}, we distill them to one step, following~\citet{salimans2022progressive}.
\paragraph{Feature vector $\bsh_1$.}
We exploit the output of the first residual block of the teacher ResNet.
\paragraph{Temperature Distribution $p_\mathrm{temp}$.}
The probability density of the temperature distribution $p_\mathrm{temp}$ follows the $\mathrm{Beta}$ distribution as follows:
$T=2(1+0.2\alpha),\quad\alpha\sim \mathrm{Beta}(\,\cdot\,;1,5)$.

The other hyperparameters are altered across the datasets and they are shown in~\cref{tab:hp}.
In addition, we also report the full list of hyperparameters used in training the baseline ensemble teacher networks in~\cref{tab:hp_ensemble}.
\begin{table}[!ht]
    \centering
\newcommand{\metricrule}{\cmidrule(lr){2-3} \cmidrule(lr){4-8}}
\newcommand{\modelrule}{\cmidrule(lr){1-8}}
    \renewcommand{\arraystretch}{1.18}
    \caption{
        The hyperparameter settings used to learn the~\gls{dbn} network.
    }
    \vspace{0.5mm}
    \resizebox{0.9\textwidth}{!}{
        \begin{tabular}{lrrrr}
            \toprule
            Dataset & CIFAR10 & CIFAR100 & TinyImageNet & ImageNet64
            \\
            \midrule
            \#Params of Teacher & 1,860,986 & 7,460,708 & 21,798,504 & 21,798,504 \\
            \#Params of Score & 395,543 & 1,547,617 & 3,186,117 & 8,278,222 \\
            Batch Size & 128 & 128 & 128 & 256 \\
            Epochs & 800 & 800 & 250 & 250 \\
            Epochs (distill) & 50 & 50 & 50 & 50 \\
            Learning Rate & 0.00025 & 0.00025 & 0.0005 & 0.0005 \\
            Learning Rate (distill) & 0.000025 & 0.000025 & 0.00005 & 0.00005 \\
            Mixup $\alpha$ & 0.4 & 0.4 & 0.4 & 0.0 \\
            $\beta_t$ & 0.0001, $t\in[0,1]$& 0.0001, $t\in[0,1]$ & 0.001, $t\in[0,1]$& 0.005, $t\in[0,1]$
            \hspace{3mm}
            \\
            \bottomrule
        \end{tabular}
    }
    \label{tab:hp}
\end{table}

\begin{table}[!ht]
\centering
\newcommand{\metricrule}{\cmidrule(lr){2-3} \cmidrule(lr){4-8}}
\newcommand{\modelrule}{\cmidrule(lr){1-8}}
\renewcommand{\arraystretch}{1.18}
\caption{
The hyperparameter settings used to learn the baseline ensemble models.
}
\vspace{0.5mm}
\resizebox{0.9\textwidth}{!}{
    \begin{tabular}{lrrrr}
        \toprule
        Dataset & CIFAR10 & CIFAR100 & TinyImageNet & ImageNet64
        \\
        \midrule
        \#Params & 1,860,986 & 7,460,708 & 21,798,504 & 21,798,504 \\
        Batch Size & 256 & 128 & 128 & 256 \\
        Epochs & 200 & 200 & 200 & 250 \\
        Learning Rate & 0.1 & 0.1 & 0.1 & 0.1 \\
        Cosine Decay Scheduling & Yes & Yes & Yes & Yes \\
        Weight Decay & 0.001 & 0.0005 & 0.0005 & 0.0001 \\
        Warmup Steps (Linear) & 0 & 5 & 5 & 5 \\
        Initial Learning Rate & 0.1 & 0.001 & 0.001 & 0.001\\
        \bottomrule
    \end{tabular}
}
\label{tab:hp_ensemble}
\end{table}

\newpage
\section{Generation Quality of Diffusion Bridge}
\label{app:sec:transitions}

\begin{figure*}[h]
    \label{fig:extra-confidences}
    \centering
    \begin{minipage}{0.98\textwidth}
    \includegraphics[width=\textwidth]{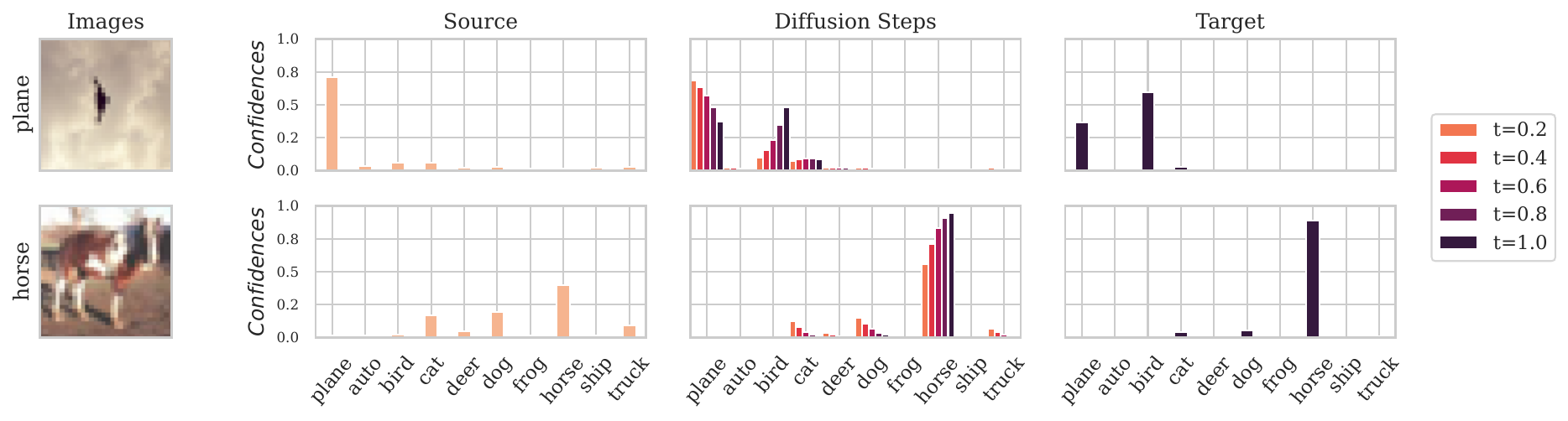}
    \end{minipage}
    \hfill
    \begin{minipage}{0.98\textwidth}
    \includegraphics[width=\textwidth]{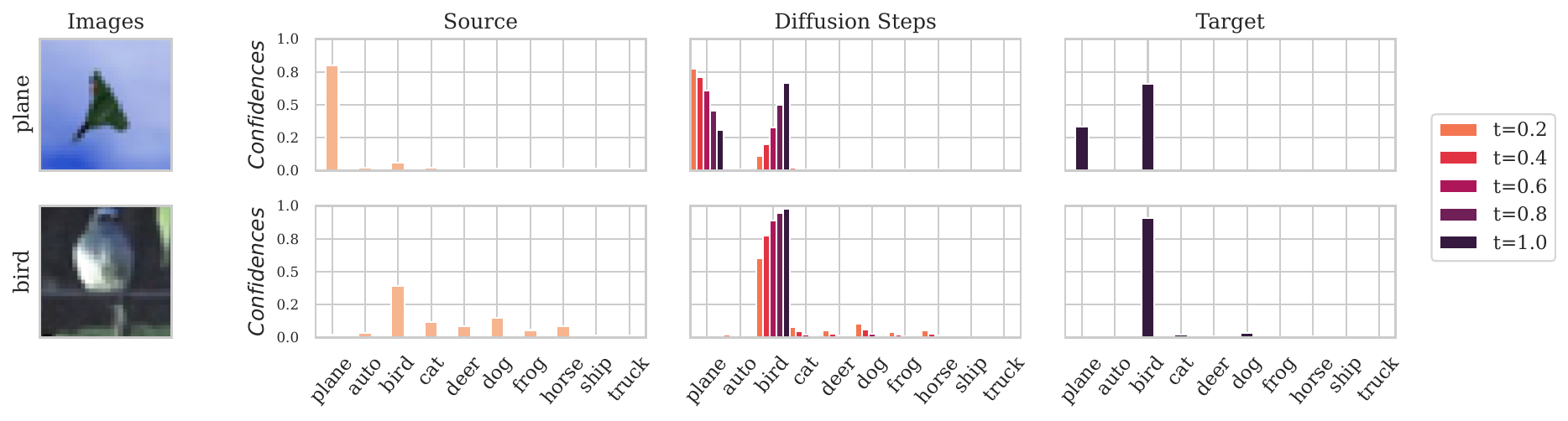}
    \end{minipage}
    \hfill
    \begin{minipage}{0.98\textwidth}
    \includegraphics[width=\textwidth]{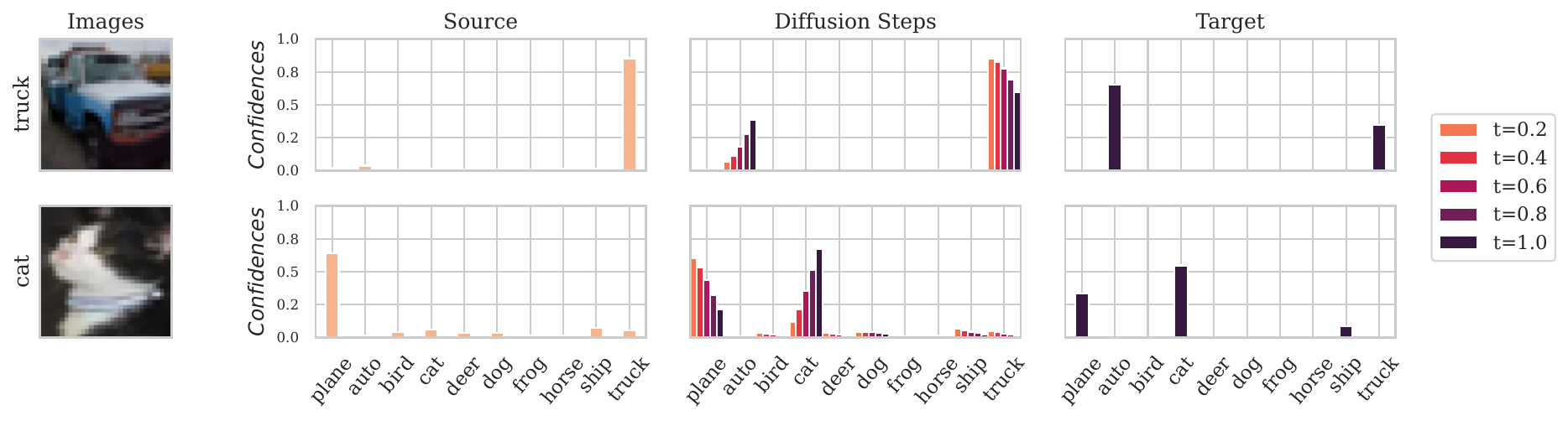}
    \end{minipage}
    \hfill
    \begin{minipage}{0.98\textwidth}
    \includegraphics[width=\textwidth]{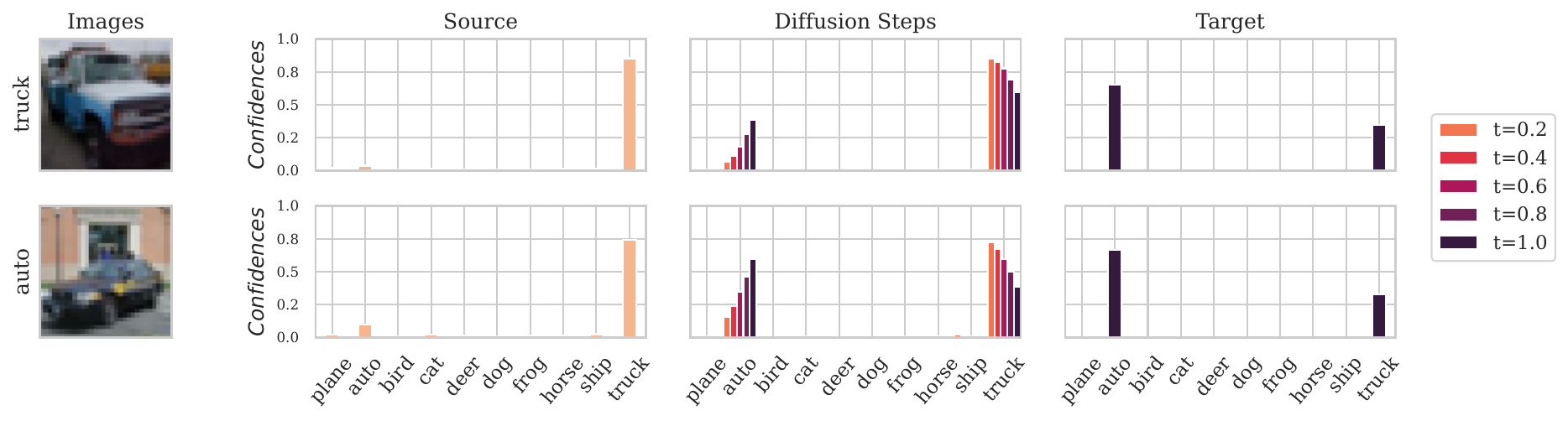}
    \end{minipage}
    \hfill
    \begin{minipage}{0.98\textwidth}
    \includegraphics[width=\textwidth]{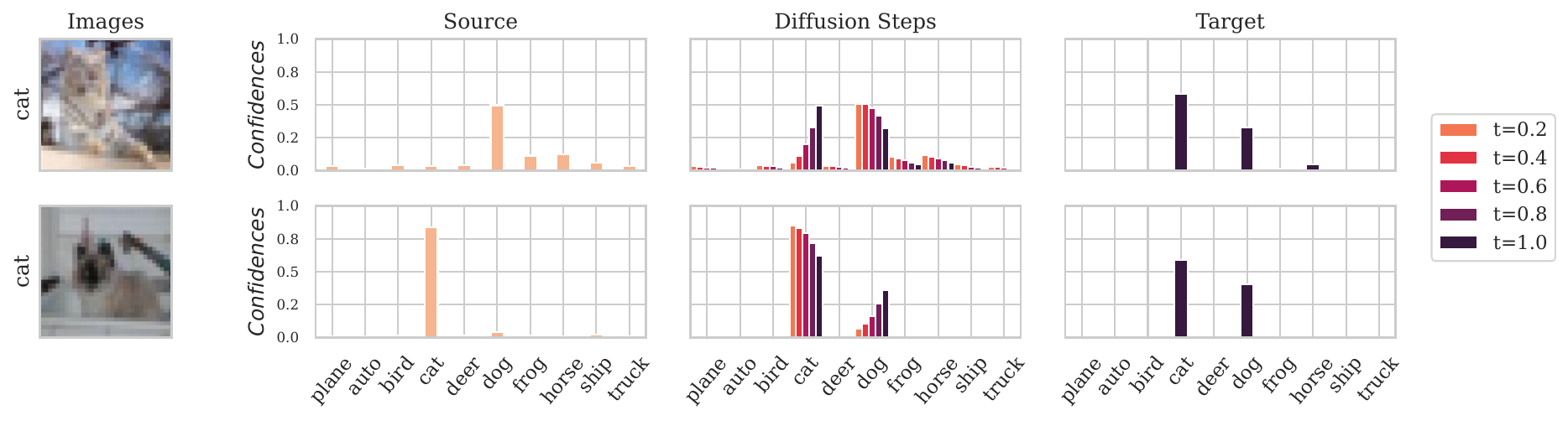}
    \end{minipage}
    \caption{Confidences from the source model (first column), from the ensemble model (third column), and from the diffusion bridge (middle column) for the given images in the CIFAR-10 dataset. The middle column illustrates a transition of the confidence from the source model to the target during the diffusion process.}
\end{figure*}
\section{Additional Experiments}
\label{app:sec:experiment}

\subsection{Results on ImageNet}

Because of the limit of the training budget, we conducted this on the downscaled 
 datasets and obtained the following result. We also attached the results of BN in the baseline ImageNet and compared the DBN and BN with the baseline DEs independently. Then we observed that the performance of our DBN method surpasses DE-2 in terms of DEE and accuracy (ACC) and it is close to that of DE-3 with only one diffusion bridge, while BN achieves the same level of performance with two bridges.

 \begin{table*}[h]
    \newcommand{\metricrule}{\cmidrule(lr){2-3} \cmidrule(lr){4-6}}
    \newcommand{\modelrule}{\cmidrule(lr){1-8}}
    \renewcommand{\arraystretch}{1.18}
    \centering
    \caption{
        Performance on ImageNet dataset. cNLL stands for NLL calibrated with an optimal temperature.
    }
    \vspace{-2mm}
    \resizebox{0.8\textwidth}{!}{
        \begin{tabular}{lrrlll}
            \\
            \toprule
            Model                 & \dmet{FLOPs} & \dmet{\#Params} & \umet{ACC}          & \dmet{cNLL}           & \dmet{DEE}           \\
            
            \midrule
            
            ResNet (\gls{de}-1)   & \relv{1.000} & \relv{1.000}    & \tv[n]{66.47}{} & \tv[n]{1.3960}{} & \tv[n]{1.000}{}      \\

            \metricrule
            
            \quad +1 \gls{dbn} (\gls{de}-3)  & \relv{1.244} & \relv{1.380}    & \tv[b]{69.79}{} & \tv[b]{1.2140}{} & \tv[b]{2.702}{}      \\

            \gls{de}-2            & \relv{2.000} & \relv{2.000}    & \tv[n]{69.56}{} & \tv[n]{1.2542}{} & \tv[n]{2.000}{} \\
            \gls{de}-3            & \relv{3.000} & \relv{3.000}    & \tv[b]{70.82}{} & \tv[b]{1.1969}{} & \tv[b]{3.000}{} \\
            \bottomrule
        \end{tabular}
    }
    \label{tab:i1000}
\end{table*}

\subsection{Out-of-distribution Performance}

We evaluated our method on the widely used CIFAR-10-C \citep{hendrycks2019benchmarking}, which is the dataset of common corruptions on the CIFAR-10 dataset. and compared it to the existing BN method as in Table 1. Achieving superior performance in the CIFAR-10-C dataset implies that the model is robust with respect to the out-of-distribution distribution, tilted by a variety of distortions and corruptions in the in-distribution dataset (in this case, CIFAR-10).

\begin{table*}[h]
    \newcommand{\metricrule}{\cmidrule(lr){1-1} \cmidrule(lr){2-4}}
    \newcommand{\modelrule}{\cmidrule(lr){1-8}}
    \renewcommand{\arraystretch}{1.18}
    \centering
    \caption{
        Test results on CIFAR-10-C. cNLL stands for NLL calibrated with an optimal temperature.
    }
    \vspace{-2mm}
    \resizebox{0.6\textwidth}{!}{
        \begin{tabular}{llll}
            \\
            \toprule
            Model                 & \umet{ACC}          & \dmet{NLL}           & \dmet{cNLL}           \\
            
            \midrule
            
            ResNet (\gls{de}-1)   & \tv[n]{86.95}{} & \tv[n]{0.5158}{} & \tv[n]{0.4477}{}      \\

            \metricrule
            \quad +2 \gls{bn}\textsubscript{medium} (\gls{de}-3)  & \tv[n]{87.80}{} & \tv[b]{0.3740}{} & \tv[n]{0.3749}{}      \\
            
            \quad +1 \gls{dbn} (\gls{de}-3)  & \tv[b]{88.91}{} & \tv[n]{0.3759}{} & \tv[b]{0.3682}{}      \\

            \gls{de}-2            & \tv[n]{88.85}{} & \tv[n]{0.3860}{} & \tv[n]{0.3693}{} \\
            \gls{de}-3            & \tv[b]{89.47}{} & \tv[b]{0.3459}{} & \tv[b]{0.3410}{} \\
            \bottomrule
        \end{tabular}
    }
    \label{tab:ood}
\end{table*}

\newpage
\section{Depthwise Separable Convolution}
\label{app:sec:depthwise}

Depthwise separable convolution is a method to modify conventional convolutional layer, which include a convolution operation between every input channel and the convolution filters.
Precisely, let the number of the input feature to have $H\times W$ spatial size with $C_\text{in}$ channels, and the convolutional layer with $h\times w$ receptive field with $C_\text{out}$ filters.
For simplicity, we take full padding and do not allow strides,and biases.
Then the number of parameters and the~\gls{flops} of the convolutional layer is $h\times w\times C_\text{in} \times C_\text{out}$ and $H\times h\times W\times w\times C_\text{in}\times C_\text{out}$, respectively.

Compared to standard convolution, the depthwise separable convolution consists of two parts: \emph{separable} convolution and \emph{point-wise} convolution.
First, \emph{separable} convolution is the convolution with $h\times w$ receptive fields and $C_\text{in}$ filters, taking the number of groups same as the input feature ($C_\text{in}$).
Then, each filter is convolved by the corresponding filters, followed by $H\times W\times C_\text{in}$ intermediate features.
The number of parameters and the~\gls{flops} of the separable convolution is $h\times w\times C_\text{in}$ and $H\times h\times W\times w\times C_\text{in}$, respectively.
Then the second part of the depthwise separable convolution is the \emph{point-wise} convolution, which is a $1\times 1$ convolution with $C_\text{in}$ input and $C_\text{out}$ output filter sizes.
The number of parameters and the~\gls{flops} of the point-wise convolution is $C_\text{in}\times C_\text{out}$ and $H\times W\times C_\text{in}\times C_\text{out}$, respectively.
We do not consider biases for evaluating the complexity of the convolution layers.

\begin{table}[!ht]
\newcommand{\metricrule}{\cmidrule(lr){2-3} \cmidrule(lr){4-8}}
    \newcommand{\modelrule}{\cmidrule(lr){1-8}}
    \renewcommand{\arraystretch}{1.18}
    \centering
    \caption{
        The complexity measures of the standard convolution and depthwise separable convolution.
    }
    \vspace{0.5mm}
    \resizebox{0.9\textwidth}{!}{
        \begin{tabular}{lcc}
            \toprule
            Model & \# Parameters & \# FLOPs      \\
            \midrule
            Standard &
            $h\times w\times C_\text{in}\times C_\text{out}$ &
            $H\times h\times W\times w\times C_\text{in}\times C_\text{out}$
            \\
            Depthwise separable &
            $(h\times w + C_\text{out}) \times C_\text{in}$ &
            $(h\times w + C_\text{out})\times H\times W\times C_\text{in}$
            \\
            D-S / Standard &
            $\dfrac{1}{C_\text{out}} + \dfrac{1}{hw}$ &
            $\dfrac{1}{C_\text{out}} + \dfrac{1}{hw}$\\
            \bottomrule
        \end{tabular}
    }
    \label{tab:depthwise_separable_convolution}
\end{table}

\end{document}